\documentclass{article}

\usepackage{graphicx,amscd,amsmath,amssymb,verbatim}
\usepackage[dvips]{hyperref}
\usepackage[TS1,OT1,T1]{fontenc}

\begin{document}

\title{Self-Organised Factorial Encoding of a Toroidal Manifold
\footnote{Submitted to Neural Computation on 18 May 1998.
Manuscript no. 1810. It was not accepted for publication,
but it underpins several subsequently published papers.}}
\author{Stephen Luttrell}
\maketitle

\noindent {\bfseries Abstract:} It is shown analytically how a neural network can be used optimally to encode
input data that is derived from a toroidal manifold. The case of a 2-layer network is considered, where the
output is assumed to be a set of discrete neural firing events. The network objective function measures the
average Euclidean error that occurs when the network attempts to reconstruct its input from its output. This
optimisation problem is solved analytically for a toroidal input manifold, and two types of solution are
obtained: a joint encoder in which the network acts as a soft vector quantiser, and a factorial encoder in which
the network acts as a pair of soft vector quantisers (one for each of the circular subspaces of the torus). The
factorial encoder is favoured for small network sizes when the number of observed firing events is large. Such
self-organised factorial encoding may be used to restrict the size of network that is required to perform a
given encoding task, and will decompose an input manifold into its constituent submanifolds.

\section{Introduction}

\label{Sect:Introduction}The purpose of this paper is to show analytically how
a neural network can be used to optimally encode input data that is derived
from a toroidal manifold. For simplicity, only the case of a 2-layer network
is considered, and an objective function is defined \cite{Ref:D1D2} that
measures the average ability of the network to reconstruct the state of its
input layer from the state of its output layer.
The optimum network parameter values must then
minimise this objective function. In this paper the output state is chosen to
be the vector of locations of a finite number of the neural firing events that
arise when an input vector is presented to the network, and, in the limit of a
single firing event, this reduces to a winner-take-all encoder network.

If the input vector is obtained from an arbitrary input probability density
function (PDF), then the network would have to be optimised numerically, and a
simple interpretation of its optimal parameters would not then be guaranteed.
On the other hand, if the input PDF is constrained to have a simple enough
form, then an analytic optimisation guarantees that the results can be
interpreted. Because the purpose of this paper is mainly to interpret the
nature of the optimal solution(s) that arise from the interplay between the
input PDF and the network objective function, an analytic rather than a
numerical approach will be used.

The detailed form of the optimum network parameters depends on the chosen
input PDF, and, for simplicity, the input PDF will be chosen to define a
curved manifold which is uniformly populated by all of the allowed input
vectors. The shape of this manifold then determines the type of optimum
solution that the network adopts. For instance, a 1-dimensional linear
manifold with a uniform distribution of input vectors leads to an optimum
solution in which each neuron fires only if the input lies within a small
range of values, so the network behaves as a soft scalar quantiser.
This result generalises to higher dimensional linear
manifolds, where the network behaves as a soft vector quantiser. A more
interesting type of optimum solution can occur when the manifold is curved.
For instance, a circular manifold (which is a 1-dimensional manifold embedded
in a 2-dimensional space) leads to an optimum solution that is analogous to
the soft scalar quantiser obtained with a 1-dimensional linear manifold, but a
toroidal manifold (which is a 2-dimensional manifold embedded in a
4-dimensional space)\ does not necessarily lead to an optimum solution that is
analogous to the soft vector quantiser obtained with a 2-dimensional linear manifold.

For a 2-dimensional toroidal manifold, it is possible for the optimum solution
to be constructed out of a pair of soft scalar quantisers, each of which
encodes only one of the two circular manifolds that form the toroidal
manifold. This is called a factorial encoder (because it breaks the input into
its constituent factors, which it then encodes), as opposed to a joint encoder
(which directly encodes the input, without first breaking it into its
constituent factors). Because a factorial encoder splits up the overall
encoding problem into a number of smaller encoding problems, which it then
tackles in parallel, it requires fewer neurons than a joint encoder would have
needed for the same encoding problem.

\begin{figure}[h]
\begin{center}
\includegraphics{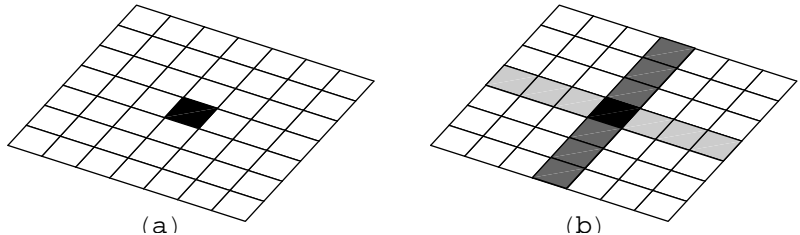}
\caption{\textit{Diagram (a)\ shows the encoding cells for joint encoding of a
2-dimensional linear manifold; a typical encoding cell is shaded. Diagram
(b)\ shows the corresponding encoding cells for a factorial encoder; typical
encoding cells for each of the two factors and their intersection are shaded.
The distortion that would result from only one of the two factors is large,
because the encoding cell is a long thin rectangular region.}}%
\label{Fig:EncodeLinear}%
\end{center}
\end{figure}

\begin{figure}[h]
\begin{center}
\includegraphics{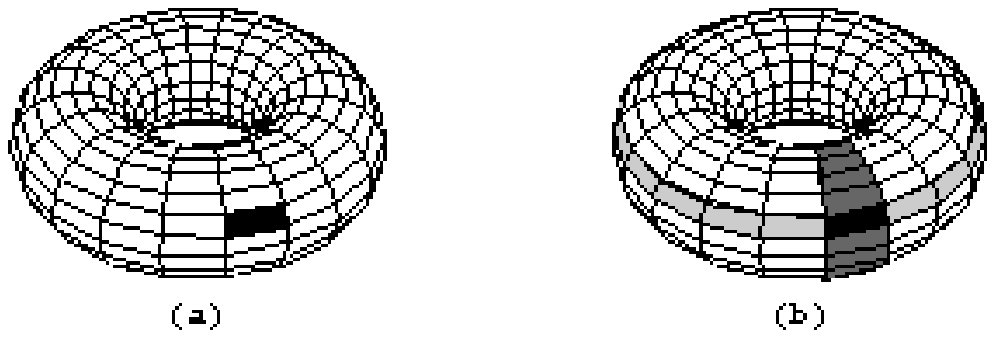}
\caption{\textit{Diagram (a) shows the encoding cells for joint encoding of a 2-dimensional toroidal manifold; a
typical encoding cell is shaded. Diagram (b) shows the corresponding encoding cells for a factorial encoder;
typical encoding cells for each of the two factors and their intersection are shaded. The distortion that would
result from only one of the two factors is not as large as in the case of the corresponding linear manifold,
because the long thin rectangular encoding cells are now wrapped round into loops, thus reducing the average
separation (in the Euclidean sense) of points within each
encoding cell.}}%
\label{Fig:EncodeCurved}%
\end{center}
\end{figure}

For the type of network objective function that is discussed in this paper,
factorial encoding does not occur with linear manifolds. This is because the
random nature of the neural firing events does not guarantee that at least one
such event occurs in each of the soft scalar quantisers in a factorial
encoder, and, for a linear manifold, this leads to a much larger average
reconstruction error if a factorial encoder is used than if a joint encoder is
used. This effect is summarised in figure \ref{Fig:EncodeLinear} for a linear
manifold, and in figure \ref{Fig:EncodeCurved} for a toroidal manifold.
Henceforth, only the toroidal case will be discussed, because it is a curved
manifold which thus has interesting factorial encoding properties, whereas a
linear manifold would not.

In figure \ref{Fig:EncodeCurved}(a) the torus is overlaid with a $20\times20
$\ toroidal lattice, and a typical joint encoding cell is highlighted (this
would use a total of $400=20\times20$\ neurons). Figure \ref{Fig:EncodeCurved}%
(a) makes clear why such encoding is described as ``joint'', because the
response of each neuron depends on the values of both dimensions of the input.
The neural network implementation of this type of joint encoder would have
connections from each output neuron to all of the input neurons.

In figure \ref{Fig:EncodeCurved}(b) the torus is overlaid with a $20\times20
$\ toroidal lattice, and a typical pair of intersecting factorial encoding
cells is highlighted (this would use a total of $40=20+20$\ neurons). Figure
\ref{Fig:EncodeCurved}(b) makes clear why such encoding is described as
``factorial'', because the response of each neuron depends on only one of the
dimensions of the input, or, in other words, on only one factor that
parameterises the input space. The neural network implementation of this type
of factorial encoder would have connections from each output neuron to only
half of the input neurons. In figure \ref{Fig:EncodeCurved}(b) an accurate
encoding is obtained by a process that is akin to triangulation, in which the
intersection between the 2 orthogonal encoding cells defines a region of the
2-torus that is equivalent to the corresponding joint encoding cell in figure
\ref{Fig:EncodeCurved}(a).

For a toroidal input manifold it turns out that there is an upper limit to the
number of neurons that can be used if a factorial encoder is to have a smaller
average reconstruction error than the corresponding joint encoder. This limit
is smaller than the number of neurons that are used in figure
\ref{Fig:EncodeCurved}(b), so that diagram should not be interpreted too literally.

\subsection{Vector Quantisers}

The existing literature on the simplest type of encoder (i.e. the vector
quantiser (VQ)) includes the following examples:

\begin{enumerate}
\item \label{VQ:Standard}A standard VQ, in which the input space is
partitioned into a number of non-overlapping encoding cells, which is also
known as an LBG vector quantiser (after the initials of the authors of
\cite{Ref:LBG}). In operation, all of the input vectors that lie closest (in
the Euclidean sense) to a given code vector are assigned the same code index
(which thus defines an encoding cell), and the approximate reconstruction of
these inputs is then the centroid of the encoding cell. This type of VQ can be
viewed as a single-layer winner-take-all (WTA) neural network.

\item
\sloppypar

A topographic VQ (TVQ), in which the code indices and encoding cells are
arranged so that code indices that differ by a small amount are assigned to
encoding cells that are close to each other (in the Euclidean sense). This
topographic property automatically emerges if a VQ is optimised for encoding
input vectors to be transmitted along a noisy communication channel
\cite{Ref:Kumazawa, Ref:Farvardin, Ref:TVQ, Ref:Burger}. The Kohonen
topographic mapping network \cite{Ref:Kohonen} is an approximation to this
type of encoder, as was explained in \cite{Ref:TVQ}. The TVQ may be
generalised to a soft TVQ (STVQ) in which each code index is chosen
probabilistically in response to the corresponding input
vector\ \cite{Ref:FMC, Ref:Graepel}.

\item \label{VQ:Multiple}Simultaneously use more than one standard VQ, with
each VQ encoding only a subspace of the input (see for example \cite{Ref:WTA}%
); in effect, more than one code index is used to encode the input vector. By
this means, a high-dimensional space can be split up into a number of lower
dimensional pieces. This type of VQ is equivalent to multiple single-layer WTA
neural network modules, each of which operates on a subspace of the input.
This is an example of a factorial encoder, in which the input is split into a
number of separate parts, or factors.

\item  The simultaneous use of multiple VQs can be extended to a tree-like
network of VQs \cite{Ref:ACE}. This type of VQ is equivalent to multiple
single layer WTA neural network modules which are connected together in a
tree-like network of modules.
\end{enumerate}

For simplicity, only the case of a 2-layer network (i.e. an input and an
output layer) will be considered, but otherwise the network will be obliged to
learn how to make use of all of its neurons. The simplest encoder which has
all of the required behaviour, and which includes the above 2-layer examples
as special cases, is one in which the neurons fire discretely in response to
the input, and, after a finite number of firing events has occurred, the input
is then reconstructed as accurately as possible (in the Euclidean sense). In
the special case where only a single firing event is observed, this reduces to
a standard LBG vector quantiser that was discussed in case \ref{VQ:Standard}
above. In the more general case, where a finite number of firing events is
observed, this can lead to factorial encoder networks of the type that was
discussed in case \ref{VQ:Multiple} above.

\subsection{Curved Manifolds}

\begin{figure}
[h]
\begin{center}
\includegraphics{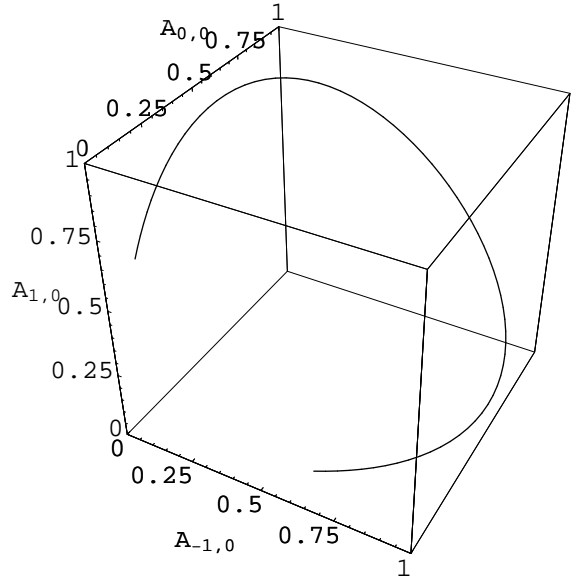}
\caption{\textit{Manifold formed when the 1-dimensional image of a target (a Gaussian profile with a half-width
of one pixel) is moved around. Only the projection }$A_{i,j}$\textit{\ onto the pixels at }$\left(  i,j\right)
=\left(  -1,0\right)  $\textit{, }$\left(  0,0\right)  $\textit{\ and
}$\left(  1,0\right)  $\textit{\ is shown.}}%
\label{Fig:Manifold1}%
\end{center}
\end{figure}

\begin{figure}[h]
\begin{center}
\includegraphics{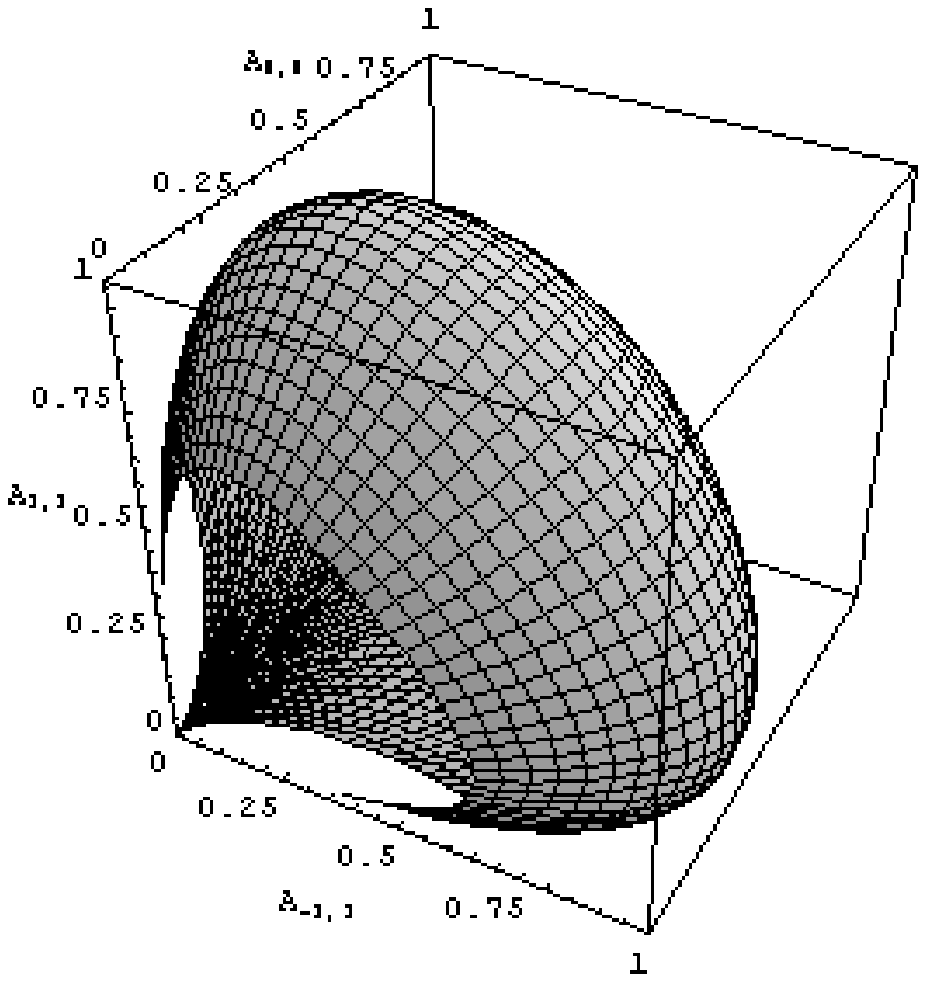}
\caption{\textit{Manifold formed when the 2-dimensional image of a target (a
Gaussian profile with half-widths of one pixel in each direction) is moved
around. Only the projection }$A_{i,j}$\textit{\ onto the pixels at }$\left(
i,j\right)  =\left(  -1,1\right)  $\textit{, }$\left(  0,0\right)
$\textit{\ and }$\left(  1,1\right)  $\textit{\ is shown.}}%
\label{Fig:Manifold2}%
\end{center}
\end{figure}

The purpose of this paper is to derive optimal ways of encoding data using
neural networks in which multiple firing events are observed, and to show that
factorial encoder networks can be optimal when the input data lies on a curved
manifold. In order to get a feel for how curved manifolds arise in image data,
consider the examples shown in figure \ref{Fig:Manifold1} and figure
\ref{Fig:Manifold2}, which show the manifold generated by a single target
(figure \ref{Fig:Manifold1}) and by a pair of targets (figure
\ref{Fig:Manifold2}), when projected onto three neighbouring pixels (i.e. the
locus of the 3-vector formed from these pixel values is plotted as the
target(s) move around).

Clearly, these image manifolds are curved, and the curvature gets greater the
narrower the Gaussian profiles used to generate the target images become.

It is not at all obvious how best to encode vectors that lie on such
manifolds. For instance, one might try to tile the manifold with a large
number of small encoding cells obtained from some variant of a VQ, or one
might try to project the manifold onto a basis obtained from some variant of
principal components analysis (PCA). In fact, these two examples are both
special cases of the approach that is advocated in this paper; a VQ
corresponds to a single firing event, whereas PCA corresponds to an infinite
number of firing events.

The problem of optimally encoding data that is derived from a general curved
manifold requires a numerical solution. However, in order to develop our
understanding, it is best to start with an analytically tractable example
based on a simple curved manifold, which is carefully selected to preserve the
essential features of more general curved manifolds. With this in mind, the
most important feature to preserve in the analytic example is curvature. A
circle is the simplest 1-dimensional curved manifold, which may then be used
to construct higher dimensional toroidal manifolds. For instance, a pair of
circles may be used to construct the 2-dimensional toroidal manifold shown in
figure \ref{Fig:EncodeCurved}. It turns out that, if a toroidal manifold is
used, then the network objective function can be analytically minimised to
yield results that exhibit interesting joint encoder and factorial encoder properties.

\subsection{Structure of this Paper}

In section \ref{Sect:BasicTheory} the basic theoretical framework is
introduced, from which some expressions are derived for optimising a network
which is trained on data from a toroidal input manifold. In section
\ref{Sect:CircularManifold} the detailed results for encoding a circular input
manifold are given (which are trivially related to the corresponding results
for the case of joint encoding of a 2-torus), and in section
\ref{Sect:ToroidalManifold} these results are extended to the case of
factorial encoding of a 2-torus. The results for joint encoding and factorial
encoding are compared in section \ref{Sect:JointVsFactorial}. Some useful
asymptotic approximations are discussed in section \ref{Sect:Asymptotic}, and
a useful approximation to the optimal network is discussed in section
\ref{Sect:ISO}.

The main steps in the derivations are reported in the appendices to this
paper, and in several cases there is a considerable amount of algebra
involved, which was done using algebraic manipulator software
\cite{Ref:Mathematica}.

\section{Basic Theoretical Framework}

\label{Sect:BasicTheory}The encoder model that is assumed throughout this
paper is a 2-layer network of neurons. The state of the input layer is denoted
as an input vector $\mathbf{x}$ (which is assumed in this paper to be a
continuous activity pattern), and the state of the output layer is denoted as
the output vector $\mathbf{y}$ (which is assumed in this paper to be a
discrete pattern of firing events). The information content of the output
state $\mathbf{y}$ may be used to draw inferences about the input state
$\mathbf{x}$. This can be formalised by using Bayes' theorem in the form
\begin{equation}
\Pr\left(  \mathbf{x}|\mathbf{y}\right)  =\frac{\Pr\left(  \mathbf{y|x}%
\right)  \Pr\left(  \mathbf{x}\right)  }{\int d\mathbf{x}^{\prime}\Pr\left(
\mathbf{y|x}^{\prime}\right)  \Pr\left(  \mathbf{x}^{\prime}\right)
}\label{Eq:Bayes}%
\end{equation}
where the PDF $\Pr\left(  \mathbf{x}|\mathbf{y}\right)  $ of the input
$\mathbf{x}$ given that the output $\mathbf{y}$ is known (i.e. the generative
model) is completely determined by two quantities: the likelihood $\Pr\left(
\mathbf{y}|\mathbf{x}\right)  $ that output $\mathbf{y}$ occurs when input
$\mathbf{x}$ is present (i.e. the recognition model), and the prior PDF
$\Pr\left(  \mathbf{x}\right)  $ that input $\mathbf{x}$ could occur
irrespective of whether $\mathbf{y}$ is being observed. However, for all but
the most trivial situations, if the functional form of $\Pr\left(
\mathbf{y|x}\right)  $ is simple then the functional form of $\Pr\left(
\mathbf{x}|\mathbf{y}\right)  $ is complicated (or vice versa, with the roles
of $\Pr\left(  \mathbf{y|x}\right)  $ and $\Pr\left(  \mathbf{x}%
|\mathbf{y}\right)  $ interchanged). In other words, if the recognition and
generative models are strictly related by Bayes' theorem, then difficulties
inevitably arise in analytic and numerical calculations.

A possible way around this problem is to use a network objective function
$D_{0}$ that has a simple functional form for the $\Pr\left(  \mathbf{y|x}%
\right)  $, but has an approximation to the ideal $\Pr\left(  \mathbf{x}%
|\mathbf{y}\right)  $ implied by Bayes' theorem (or vice versa). A convenient
choice is
\begin{align}
D_{0}  & \equiv-\int d\mathbf{x}\sum_{\mathbf{y}}\Pr\left(  \mathbf{x}%
,\mathbf{y}\right)  \log Q\left(  \mathbf{x},\mathbf{y}\right) \nonumber\\
& =-\int d\mathbf{x}\Pr\left(  \mathbf{x}\right)  \sum_{\mathbf{y}}\Pr\left(
\mathbf{y}|\mathbf{x}\right)  \log Q\left(  \mathbf{x}|\mathbf{y}\right)
-\sum_{\mathbf{y}}\Pr\left(  \mathbf{y}\right)  \log Q\left(  \mathbf{y}%
\right) \label{Eq:EncodeBits}%
\end{align}

\sloppypar

$\Pr\left(  \mathbf{x},\mathbf{y}\right)  $ is a joint probability that satisfies $\Pr\left(
\mathbf{x},\mathbf{y}\right)  =$ $\Pr\left(
\mathbf{y|x}\right)  \Pr\left(  \mathbf{x}\right)  =\Pr\left(  \mathbf{x}%
|\mathbf{y}\right)  \Pr\left(  \mathbf{y}\right)  $ (i.e. Bayes' theorem
holds), $Q\left(  \mathbf{x},\mathbf{y}\right)  $ is an approximation to
$\Pr\left(  \mathbf{x},\mathbf{y}\right)  $ that satisfies the corresponding
relationships $Q\left(  \mathbf{x},\mathbf{y}\right)  =$ $Q\left(
\mathbf{y|x}\right)  Q\left(  \mathbf{x}\right)  =Q\left(  \mathbf{x}%
|\mathbf{y}\right)  Q\left(  \mathbf{y}\right)  $, $\int d\mathbf{x}\Pr\left(
\mathbf{x}\right)  \left(  \cdots\right)  $ integrates over all the possible
states of the input layer, $\sum_{\mathbf{y}}\Pr\left(  \mathbf{y}%
|\mathbf{x}\right)  \left(  \cdots\right)  $ sums over all the possible states
of the output layer given that the state of the input layer is known, and
$\sum_{\mathbf{y}}\Pr\left(  \mathbf{y}\right)  \left(  \cdots\right)  $ sums
over all the possible states of the output layer.

\sloppypar

The objective function $D_{0}$ measures the average number of bits required
when the approximate joint probability $Q\left(  \mathbf{x},\mathbf{y}\right)
$ is used as a reference to encode each pair $\left(  \mathbf{x}%
,\mathbf{y}\right)  $ drawn randomly from the true joint probability
$\Pr\left(  \mathbf{x},\mathbf{y}\right)  $ \cite{Ref:Information}, so $D_{0}$
belongs to the class of minimum description length (MDL) objective functions
\cite{Ref:Rissanen}.
Strictly
speaking, the number of bits depends on the accuracy with which the
continuous-valued $\mathbf{x}$ is measured. However, this refinement is
omitted from equation \ref{Eq:EncodeBits} because it does not affect the
results in this paper, provided that the size of the quantisation cells into
which $\mathbf{x}$ is binned is much smaller than the scale on which
$\Pr\left(  \mathbf{x}|\mathbf{y}\right)  $ and $Q\left(  \mathbf{x}%
|\mathbf{y}\right)  $ fluctuate.

The objective function $D_{0}$ can be simplified if $Q\left(  \mathbf{x}%
,\mathbf{y}\right)  $ is assumed to have the following properties
\begin{align}
Q\left(  \mathbf{y}\right)   & =\text{constant}\nonumber\\
Q\left(  \mathbf{x}|\mathbf{y}\right)   & =\frac{1}{\left(  \sqrt{2\pi}%
\sigma\right)  ^{\dim\mathbf{x}}}\exp\left(  -\frac{\left\|  \mathbf{x}%
-\mathbf{x}^{\prime}\left(  \mathbf{y}\right)  \right\|  ^{2}}{2\sigma^{2}%
}\right) \label{Eq:SimplifiedQ}%
\end{align}
where the approximation $Q\left(  \mathbf{x}|\mathbf{y}\right)  $ to the true
generative model $\Pr\left(  \mathbf{x}|\mathbf{y}\right)  $ is a Gaussian
PDF, and the prior probabilities $Q\left(  \mathbf{y}\right)  $ are
constrained to all be equal. If the value of $\sigma$ is fixed, then $D_{0}$
may be replaced by the simpler, but equivalent, vector quantiser objective
function $D_{VQ}$, which is defined as
\begin{equation}
D_{VQ}\equiv\int d\mathbf{x}\Pr\left(  \mathbf{x}\right)  \sum_{\mathbf{y}}%
\Pr\left(  \mathbf{y}|\mathbf{x}\right)  \left\|  \mathbf{x}-\mathbf{x}%
^{\prime}\left(  \mathbf{y}\right)  \right\|  ^{2}\label{Eq:EncodeDistortion}%
\end{equation}
where $\sum_{\mathbf{y}}\Pr\left(  \mathbf{y}\right)  =1$ has been used to
eliminate the $\sum_{\mathbf{y}}\Pr\left(  \mathbf{y}\right)  \log Q\left(
\mathbf{y}\right)  $ term. This measures the average Euclidean distortion that
occurs when the input $\mathbf{x}$ is probabilistically encoded as
$\mathbf{y}$, and then subsequently reconstructed as $\mathbf{x}^{\prime
}\left(  \mathbf{y}\right)  $. This is a soft version of the LBG\ vector
quantiser objective function \cite{Ref:LBG}, in which $\mathbf{y}$ acts as a
code index, $\Pr\left(  \mathbf{y}|\mathbf{x}\right)  $ acts a soft encoding
prescription for probabilistically transforming $\mathbf{x}$ into $\mathbf{y}
$, and $\mathbf{x}^{\prime}\left(  \mathbf{y}\right)  $ acts as the
corresponding code vector. The optimal $\Pr\left(  \mathbf{y}|\mathbf{x}%
\right)  $ that minimises $D_{VQ}$ is deterministic (i.e. each $\mathbf{x}$ is
transformed to one, and only one, $\mathbf{y}$), so $D_{VQ}$ actually leads to
an LBG vector quantiser itself, rather than merely a probabilistic version
thereof \cite{Ref:FMC}.

Under the same assumptions (see equation \ref{Eq:SimplifiedQ}) that yielded
the expression for $D_{VQ}$, the Helmholtz machine objective function
\cite{Ref:Dayan}
would
reduce to
\begin{equation}
D_{HM}=D_{VQ}+\int d\mathbf{x}\Pr\left(  \mathbf{x}\right)  \sum_{\mathbf{y}%
}\Pr\left(  \mathbf{y}|\mathbf{x}\right)  \log\Pr\left(  \mathbf{y}%
|\mathbf{x}\right) \label{Eq:ObjectiveHelmholtz}%
\end{equation}
where the extra term is the so-called ``bits-back'' term, which is (minus) the
entropy of the output $\mathbf{y}$ given that the input $\mathbf{x}$ is known,
then averaged over all inputs. Thus $D_{HM}$ does not directly penalise
$\Pr\left(  \mathbf{y}|\mathbf{x}\right)  $ that have a large entropy, or, in
other words, it allows the recognition model $\Pr\left(  \mathbf{y}%
|\mathbf{x}\right)  $ to be such that many output states $\mathbf{y} $ are
permitted once the input state $\mathbf{x}$ is known. This means that the
recognition models produced by a Helmholtz machine tend to be more stochastic
than they would have been had the ``bits-back'' term been omitted from
$D_{HM}$. Conversely, the objective function $D_{VQ}$ that is used in this
paper directly penalises $\Pr\left(  \mathbf{y}|\mathbf{x}\right)  $ that have
a large entropy, so the recognition models produced tend to be more
deterministic than the stochastic ones that the Helmholtz machine would
produce under equivalent circumstances. Thus using $D_{VQ}$ tends to lead to
sparse codes in which few neurons can fire, whereas using $D_{HM}$ tends to
lead to distributed codes in which many neurons can fire.

The chosen objective function has both an information theoretic interpretation
(given by $D_{0}$ in equation \ref{Eq:EncodeBits}), in which it seeks to
minimise the number of bits required to encode $\Pr\left(  \mathbf{x}%
,\mathbf{y}\right)  $, and also an encoder/decoder interpretation (given by
$D_{VQ}$ in equation \ref{Eq:EncodeDistortion}), in which it seeks to minimise
the Euclidean distortion that arises when $\mathbf{x}$ is encoded as
$\mathbf{y}$ and then subsequently reconstructed as $\mathbf{x}^{\prime
}\left(  \mathbf{y}\right)  $. Also, using $D_{VQ}$ as the network objective
function ensures backward compatibility with preexisting results (e.g.
\cite{Ref:LBG, Ref:Kohonen}).

An upper bound on the network objective function is introduced in section
\ref{Sect:ObjectiveFunction}, and the stationarity conditions which must be
satisfied for an optimal network behaviour are derived in section
\ref{Sect:StationarityConditions}. Joint encoding on a 2-torus is discussed in
section \ref{Sect:Joint}, and factorial encoding on a 2-torus is discussed in
section \ref{Sect:Factorial}.

\subsection{Objective Function}

\label{Sect:ObjectiveFunction}In order to make progress it is necessary to
make some assumptions about the network output state $\mathbf{y}$. Thus the
output layer will be assumed to consist of $M$ neurons that fire discretely in
response to the input activity pattern $\mathbf{x}$. Furthermore, $\mathbf{y}$
will be assumed to be an $n$-dimensional vector, that consists of the
observations of the locations $\left(  y_{1},y_{2},\cdots,y_{n}\right)  $ of
the first $n$ firing events that occur in response to input $\mathbf{x}$ (this
is described in detail in \cite{Ref:D1D2}). Note that the individual $y_{i}$
are scalars, but the generalisation to vector-valued $\mathbf{y}_{i}$ is straightforward.%

\sloppypar

For compatibility with results published earlier (e.g. \cite{Ref:FMC,
Ref:D1D2}), the objective function that will be used here is $D=2D_{VQ}$,
which has an upper bound $D_{1}+D_{2}$ given by (see appendix
\ref{Appendix:Objective} for a detailed derivation and discussion)
\begin{align}
D_{1}  & \equiv\frac{2}{n}\int d\mathbf{x}\Pr\left(  \mathbf{x}\right)
\sum_{y=1}^{M}\Pr\left(  y|\mathbf{x}\right)  \left\|  \mathbf{x}%
-\mathbf{x}^{\prime}\left(  y\right)  \right\|  ^{2}\nonumber\\
D_{2}  & \equiv\frac{2\left(  n-1\right)  }{n}\int d\mathbf{x}\Pr\left(
\mathbf{x}\right)  \left\|  \mathbf{x}-\sum_{y=1}^{M}\Pr\left(  y|\mathbf{x}%
\right)  \mathbf{x}^{\prime}\left(  y\right)  \right\|  ^{2}%
\label{Eq:ObjectiveD1D2}%
\end{align}
where $\Pr\left(  y|\mathbf{x}\right)  $ is the probability that neuron $y$
fires first in response to input $\mathbf{x}$, and $\mathbf{x}^{\prime}\left(
y\right)  $ is a reference vector that is used by neuron $y$ in its attempt to
approximately reconstruct the input. In the limit $n=1$ only $D_{1}$
contributes, and a standard LBG vector quantiser emerges when $D_{1}$ is
minimised. As $n\rightarrow\infty$ only $D_{2}$ contributes, and a PCA encoder
emerges when $D_{2}$ is minimised.

This upper bound $D_{1}+D_{2}$ on the objective function $D$ will be used to
derive all of the results in this paper. Its functional form, in which
$\Pr\left(  y|\mathbf{x}\right)  $ appears only quadratically (unlike in
equation \ref{Eq:ObjectiveHelmholtz} for $D_{HM}$), allows analytic results to
be readily derived.

\subsection{Stationarity Conditions}

\label{Sect:StationarityConditions}The upper bound $D_{1}+D_{2}$ (see equation
\ref{Eq:ObjectiveD1D2}) on the objective function $D=2D_{VQ}$ (see equation
\ref{Eq:EncodeDistortion}) needs to be minimised with respect to two types of
parameter: posterior probabilities $\Pr\left(  y|\mathbf{x}\right)  $ and
reference vectors $\mathbf{x}^{\prime}\left(  y\right)  $. This could be done
numerically for an arbitrary input PDF $\Pr\left(  \mathbf{x}\right)  $ by
using a gradient descent type of algorithm \cite{Ref:D1D2}, but here
$D_{1}+D_{2}$ will be analytically minimised for some carefully chosen special
cases of $\Pr\left(  \mathbf{x}\right)  $.

The stationarity condition $\frac{\partial\left(  D_{1}+D_{2}\right)
}{\partial\mathbf{x}^{\prime}\left(  y\right)  }=0$ gives (see appendix
\ref{Appendix:StationarityX})
\begin{equation}
n\int d\mathbf{x}\Pr\left(  \mathbf{x}|y\right)  \,\mathbf{x=x}^{\prime
}\left(  y\right)  +\left(  n-1\right)  \int d\mathbf{x}\Pr\left(
\mathbf{x}|y\right)  \sum_{y^{\prime}=1}^{M}\Pr\left(  y^{\prime}%
|\mathbf{x}\right)  \mathbf{x}^{\prime}\left(  y^{\prime}\right)
\label{Eq:StationaryX2}%
\end{equation}
where $\Pr\left(  y\right)  >0$ has been assumed. The $\frac{\partial\left(
D_{1}+D_{2}\right)  }{\partial\mathbf{x}^{\prime}\left(  y\right)  }=0$
stationarity condition also has the solution $\Pr\left(  y\right)  =0$, but
this solution may be discarded because $\Pr\left(  y\right)  >0$ is always the
case in practice. The right hand side of the stationarity condition in
equation \ref{Eq:StationaryX2} has two contributions: a $D_{1}$-like
contribution which is a single reference vector $\mathbf{x}^{\prime}\left(
y\right)  $, plus a $D_{2}$-like contribution which is $n-1$ times a sum of
reference vectors $\sum_{y^{\prime}=1}^{M}\left(  \int d\mathbf{x}\Pr\left(
y^{\prime}|\mathbf{x}\right)  \Pr\left(  \mathbf{x}|y\right)  \right)
\mathbf{x}^{\prime}\left(  y^{\prime}\right)  $, where the coefficient $\int
d\mathbf{x}\Pr\left(  y^{\prime}|\mathbf{x}\right)  \Pr\left(  \mathbf{x}%
|y\right)  $ accounts for the effect (at neuron $y$) of observing all pairs of
firing events $\left(  y,y^{\prime}\right)  $ for $y^{\prime}=1,2,\cdots,M$.
The sum of these two terms is $n$ times the total reference vector that is
effectively associated with neuron $y$, which is $n$ times $\int
d\mathbf{x}\Pr\left(  \mathbf{x}|y\right)  \,\mathbf{x}$ as given on the left
hand side of equation \ref{Eq:StationaryX2}.

The stationarity condition $\frac{\delta\left(  D_{1}+D_{2}\right)  }%
{\delta\log\Pr\left(  y|\mathbf{x}\right)  }=0$ gives (see appendix
\ref{Appendix:StationarityP})
\begin{equation}
\sum_{y^{\prime}=1}^{M}\left(  \Pr\left(  y^{\prime}|\mathbf{x}\right)
-\delta_{y,y^{\prime}}\right)  \,\mathbf{x}^{\prime}\left(  y^{\prime}\right)
\cdot\left(  \frac{1}{2}\mathbf{x}^{\prime}\left(  y^{\prime}\right)
-n\,\mathbf{x}+\left(  n-1\right)  \sum_{y^{\prime\prime}=1}^{M}\Pr\left(
y^{\prime\prime}|\mathbf{x}\right)  \mathbf{x}^{\prime}\left(  y^{\prime
\prime}\right)  \right)  =0\label{Eq:StationaryP}%
\end{equation}
where the constraint $\sum_{y^{\prime}=1}^{M}\Pr\left(  y^{\prime}%
|\mathbf{x}\right)  =1$ has been imposed, and $\Pr\left(  \mathbf{x}\right)
>0$ and $\Pr\left(  y|\mathbf{x}\right)  >0$ have been assumed. The
$\frac{\delta\left(  D_{1}+D_{2}\right)  }{\delta\log\Pr\left(  y|\mathbf{x}%
\right)  }=0$ stationarity condition also has two other solutions:\ either
$\Pr\left(  \mathbf{x}\right)  =0$, or $\Pr\left(  \mathbf{x}\right)  >0$ and
$\Pr\left(  y|\mathbf{x}\right)  =0$. Using the normalisation constraint
$\sum_{y=1}^{M}\Pr\left(  y|\mathbf{x}\right)  =1$, the last of these
solutions ensures that $\Pr\left(  y^{\prime}|\mathbf{x}\right)  \leq1$ for
$y^{\prime}\neq y$, and when all values of $y$ are considered the net effect
is to constrain $\Pr\left(  y|\mathbf{x}\right)  $ to the interval $0\leq
\Pr\left(  y|\mathbf{x}\right)  \leq1$, as expected.

The solutions of the stationarity condition for $\Pr\left(  y|\mathbf{x}%
\right)  $ in equation \ref{Eq:StationaryP} are piecewise linear functions of
$\mathbf{x}$.
This piecewise linear property of $\Pr\left(
y|\mathbf{x}\right)  $ (as discussed in appendix \ref{Appendix:StationarityP}%
)\ is an enormous simplification, because it means that rather than searching
the infinite dimensional space of functions $\Pr\left(  y|\mathbf{x}\right)  $
for the optimal ones that minimise $D_{1}+D_{2}$, one needs only search a
finite dimensional space of piecewise linear functions $\Pr\left(
y|\mathbf{x}\right)  $ (subject to the constraints $0\leq\Pr\left(
y|\mathbf{x}\right)  \leq1$ and $\sum_{y=1}^{M}\Pr\left(  y|\mathbf{x}\right)
=1$).

\subsection{Joint Encoding}

\label{Sect:Joint}Joint encoding, as shown in figure \ref{Fig:EncodeCurved}%
(a), is characterised by a $\Pr\left(  y|\mathbf{x}\right)  $ in which the
neurons labelled by $y$ form a discretised version of the manifold that
$\mathbf{x}$ lives on. For instance, when $\mathbf{x}$ lives on a 2-torus, so
that $\mathbf{x}=\left(  \mathbf{x}_{1},\mathbf{x}_{2}\right)  $ where
$\mathbf{x}_{1}=\left(  \cos\theta_{1},\sin\theta_{1}\right)  $ and
$\mathbf{x}_{2}=\left(  \cos\theta_{2},\sin\theta_{2}\right)  $, where
$0\leq\theta_{1}<2\pi$ and $0\leq\theta_{2}<2\pi$, the $\Pr\left(
y|\mathbf{x}\right)  $ typically behave as shown in figure
\ref{Fig:EncodeCurved}(a), where the 2-torus is tiled with encoding cells.
When $n>1$ neighbouring encoding cells overlap, so figure
\ref{Fig:EncodeCurved}(a) does not then give an accurate representation of the
encoding cells.

For joint encoding of a 2-torus, $y$ must be replaced by the pair $\left(
y_{1},y_{2}\right)  $, where the $y_{1}$ index labels one direction around the
toroidal lattice, and $y_{2}$ labels the other direction (this notation must
not be confused with the $\left(  y_{1},y_{2},\cdots,y_{n}\right)  $ notation
that was used in section \ref{Sect:ObjectiveFunction}). Thus $\Pr\left(
y|\mathbf{x}\right)  \rightarrow\Pr\left(  y_{1},y_{2}|\mathbf{x}%
_{1},\mathbf{x}_{2}\right)  $ with $1\leq y_{1}\leq\sqrt{M}$ and $1\leq
y_{2}\leq\sqrt{M}$. For simplicity, assume $\Pr\left(  \mathbf{x}%
_{1},\mathbf{x}_{2}\right)  =\Pr\left(  \mathbf{x}_{1}\right)  \Pr\left(
\mathbf{x}_{2}\right)  $, where $\Pr\left(  \mathbf{x}_{1}\right)  $ and
$\Pr\left(  \mathbf{x}_{2}\right)  $ each define a uniform PDF on the input
manifold. The following results for $D_{1}$ and $D_{2}$ may then be derived
(see appendix \ref{Appendix:Joint})
\begin{align}
D_{1}  & =\frac{4}{n}\int d\mathbf{x}_{1}\Pr\left(  \mathbf{x}_{1}\right)
\sum_{y_{1}=1}^{\sqrt{M}}\Pr\left(  y_{1}|\mathbf{x}_{1}\right)  \left\|
\mathbf{x}_{1}-\mathbf{x}_{1}^{\prime}\left(  y_{1}\right)  \right\|
^{2}\nonumber\\
D_{2}  & =\frac{4\left(  n-1\right)  }{n}\int d\mathbf{x}_{1}\Pr\left(
\mathbf{x}_{1}\right)  \left\|  \mathbf{x}_{1}-\sum_{y_{1}=1}^{\sqrt{M}}%
\Pr\left(  y_{1}|\mathbf{x}_{1}\right)  \,\mathbf{x}_{1}^{\prime}\left(
y_{1}\right)  \right\|  ^{2}\label{Eq:D1D2Joint}%
\end{align}

These results for $D_{1}$ and $D_{2}$ show that, under the simplifying
assumptions made above, the problem of optimising a joint encoder is
equivalent to the problem of optimising an encoder for $\mathbf{x}_{1}$ alone
(with the replacement $M\rightarrow\sqrt{M}$), and then multiplying the value
of $D_{1}+D_{2}$ by a factor 2 to account for $\mathbf{x}_{2}$ as well. This
illustration of the behaviour of joint encoder posterior probabilities in the
case of $\Pr\left(  y_{1},y_{2}|\mathbf{x}_{1},\mathbf{x}_{2}\right)  $ may
readily be generalised to higher dimensions.

\subsection{Factorial Encoding}

\label{Sect:Factorial}Factorial encoding, as shown in figure
\ref{Fig:EncodeCurved}(b), is characterised by a $\Pr\left(  y|\mathbf{x}%
\right)  $ in which the neurons labelled by $y$ are partitioned into a number
of subsets, each of which forms a discretised version of a subspace of the
manifold that $\mathbf{x}$ lives on. For instance, when $\mathbf{x}$ lives on
a 2-torus, and the neurons are partitioned into two equal-sized subsets, the
$\Pr\left(  y|\mathbf{x}\right)  $ typically behave as shown in figure
\ref{Fig:EncodeCurved}(b), where each of the two circular subspaces within the
2-torus is tiled with encoding cells, which overlap when $n>1$.

For factorial encoding of a 2-torus $\Pr\left(  y|\mathbf{x}\right)
=\Pr\left(  y|\mathbf{x}_{1},\mathbf{x}_{2}\right)  =\frac{1}{2}\Pr\left(
y|\mathbf{x}_{1}\right)  +\frac{1}{2}\Pr\left(  y|\mathbf{x}_{2}\right)  $,
where $\sum_{y=1}^{\frac{M}{2}}\Pr\left(  y|\mathbf{x}_{1}\right)  =1$,
$\sum_{y=\frac{M}{2}+1}^{M}\Pr\left(  y|\mathbf{x}_{2}\right)  =1$,
$\Pr\left(  y|\mathbf{x}_{1}\right)  =0$ for $\frac{M}{2}+1\leq y\leq M$, and
$\Pr\left(  y|\mathbf{x}_{2}\right)  =0$ for $1\leq y\leq\frac{M}{2}$. For
simplicity, assume $\Pr\left(  \mathbf{x}_{1},\mathbf{x}_{2}\right)
=\Pr\left(  \mathbf{x}_{1}\right)  \Pr\left(  \mathbf{x}_{2}\right)  $, where
$\Pr\left(  \mathbf{x}_{1}\right)  $ and $\Pr\left(  \mathbf{x}_{2}\right)  $
each define a uniform PDF on the input manifold. The following results for
$D_{1}$ and $D_{2}$ may then be derived (see appendix \ref{Appendix:Factorial}%
)
\begin{align}
D_{1}  & =\frac{2}{n}\left(  \int d\mathbf{x}_{1}\Pr\left(  \mathbf{x}%
_{1}\right)  \sum_{y=1}^{\frac{M}{2}}\Pr\left(  y|\mathbf{x}_{1}\right)
\left\|  \mathbf{x}_{1}-\mathbf{x}_{1}^{\prime}\left(  y\right)  \right\|
^{2}+\int d\mathbf{x}_{2}\Pr\left(  \mathbf{x}_{2}\right)  \left\|
\mathbf{x}_{2}\right\|  ^{2}\right) \nonumber\\
D_{2}  & =\frac{4\left(  n-1\right)  }{n}\int d\mathbf{x}_{1}\Pr\left(
\mathbf{x}_{1}\right)  \left\|  \mathbf{x}_{1}-\frac{1}{2}\sum_{y=1}^{\frac
{M}{2}}\Pr\left(  y|\mathbf{x}_{1}\right)  \mathbf{x}_{1}^{\prime}\left(
y\right)  \right\|  ^{2}\label{Eq:D1D2Factorial}%
\end{align}

These results for $D_{1}$ and $D_{2}$ show that, under the simplifying
assumptions made above, the problem of optimising a factorial encoder is
closely related to the problem of optimising two 1-dimensional encoders. This
illustration of the behaviour of factorial encoder posterior probabilities in
the case of $\Pr\left(  y|\mathbf{x}_{1},\mathbf{x}_{2}\right)  $ may readily
be generalised to higher dimensions.

\section{Circular Manifold}

\label{Sect:CircularManifold}The analysis of how to encode data that lives on
a curved manifold begins with the case of data that lives on a circle. In
particular, assume that the input vector $\mathbf{x}$ is uniformly distributed
on the unit circle centred on the origin, so that $\mathbf{x}$ can be
parameterised by a single angular variable $\theta$, thus
\begin{align}
\mathbf{x}  & =\left(  \cos\theta,\sin\theta\right) \nonumber\\
\int d\mathbf{x\,}\Pr\left(  \mathbf{x}\right)  \,\left(  \cdots\right)   &
=\frac{1}{2\pi}\int_{0}^{2\pi}d\theta\,\left(  \cdots\right)
\label{Eq:UnitCircle}%
\end{align}
The posterior probability $\Pr\left(  y|\mathbf{x}\right)  $ may thus be
replaced by $\Pr\left(  y|\theta\right)  $, and for purely conventional
reasons, the range of $y$ is now chosen to be $y=0,1,\cdots,M-1$ rather than
$y=1,2,\cdots,M$. The set of $M$ posterior probabilities for $y=0,1,\cdots
,M-1$ can be parameterised as
\begin{equation}
\Pr\left(  y|\theta\right)  =p\left(  \theta-\frac{2\pi y}{M}\right)
\label{Eq:PosteriorCircle}%
\end{equation}
where\ $p\left(  \theta\right)  $ is the $\theta$-dependence of the posterior
probability associated with the $y=0$ neuron. The $\theta$-dependence of
$p\left(  \theta\right)  $ must be piecewise sinusoidal (i.e. made out of
pieces that each have the functional form $a+b\cos\theta+c\sin\theta$)\ in
order to ensure that $\Pr\left(  y|\mathbf{x}\right)  $ is piecewise linear,
as is required of solutions to equation \ref{Eq:StationarityP}. Similarly, the
$M$ corresponding reference vectors can be parameterised as
\begin{equation}
\mathbf{x}^{\prime}\left(  y\right)  =r\,\left(  \cos\left(  \frac{2\pi y}%
{M}\right)  ,\sin\left(  \frac{2\pi y}{M}\right)  \right) \label{Eq:RV}%
\end{equation}
which all have length $r$, and thus form a regular $M$-sided polygon.

It turns out that, for input vectors that live on a circular manifold, optimal
joint encoding never causes more than 3 different neurons to fire in response
to a given input (i.e. no more than 3 posterior probabilities overlap in input
space). This severely limits the number of different piecewise functions that
have to be manipulated when solving the $D_{1}+D_{2} $ minimisation problem
for input vectors that live on a circle. An analogous simplification also
holds for joint and factorial encoding of a 2-torus. The case of 2 overlapping
posterior probabilities can be optimised without too much difficulty, but the
case of 3 overlapping posterior probabilities involves a prohibitively large
amount of algebra, for which it is convenient to use an algebraic manipulator
\cite{Ref:Mathematica}. The calculations turn out to be highly structured, so
the use of an algebraic manipulator could in principle be used to solve even
more complicated analytic problems.

All of the results for encoding input data that lives on a circular manifold
may be derived from the expression for $D_{1}+D_{2}$ in equation
\ref{Eq:ObjectiveD1D2} (and the corresponding stationarity conditions), with
the replacement given in equation \ref{Eq:UnitCircle} to ensure that the input
manifold corresponds to a uniform distribution of data around a unit circle,
and the functional forms given in equation \ref{Eq:PosteriorCircle} and
equation \ref{Eq:RV}.

The corresponding results for joint encoding of data that lives on a 2-torus
can be obtained directly from these results (see section \ref{Sect:Joint}).
The expression for the minimum value of $D_{1}+D_{2}$ for joint encoding a
2-torus using $\sqrt{M}\times\sqrt{M}$ neurons is obtained by making the
replacement $M\rightarrow\sqrt{M}$ in the expression for the minimum value of
$D_{1}+D_{2}$ for encoding a circle using $M$ neurons, and then multiplying
this result by 2 in order to account for both the circles that form the
2-torus (see equation \ref{Eq:D1D2Joint}).

\subsection{Two Overlapping Posterior Probabilities}

\label{Sect:CircularManifold2}
\begin{figure}
[h]
\begin{center}
\includegraphics{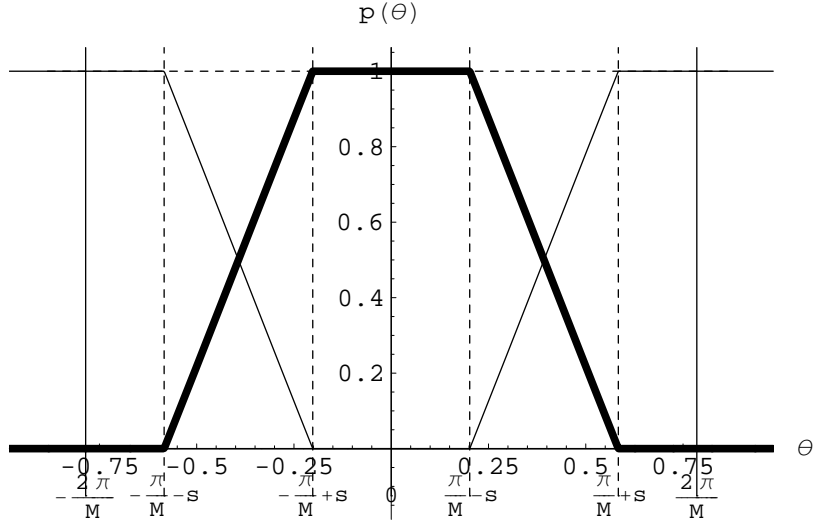}
\caption{\textit{Plot of the optimal neural posterior probability }$p\left( \theta\right)  $\textit{\ for
}$M=8$\textit{\ and }$n=2$\textit{. The
neighbouring posterior probabilities }$p\left(  \theta\pm\frac{2\pi}%
{M}\right)  $\textit{\ are also plotted. The optimal value of }$s$\textit{\ is
}$s\approx0.49\frac{\pi}{M}$\textit{. The departure of }$p\left(
\theta\right)  $\textit{\ from linearity in the interval }$\frac{\pi}{M}%
-s\leq\theta\leq\frac{\pi}{M}+s$\textit{\ is too small to be easily seen.}}%
\label{Fig:PosteriorCircle2}%
\end{center}
\end{figure}

A detailed derivation of the results reported in
this section is given in appendix \ref{Appendix:CircularManifold2}. Because
the neurons have an angular separation of $\frac{2\pi}{M}$ (see the form of
the posterior probability given in equation \ref{Eq:PosteriorCircle}), the
functional form of $p\left(  \theta\right)  $ may be defined as
\begin{equation}
p\left(  \theta\right)  =\left\{
\begin{array}
[c]{ll}%
1 & 0\leq\left|  \theta\right|  \leq\frac{\pi}{M}-s\\
f\left(  \theta\right)  & \frac{\pi}{M}-s\leq\left|  \theta\right|  \leq
\frac{\pi}{M}+s\\
0 & \left|  \theta\right|  \geq\frac{\pi}{M}+s
\end{array}
\right. \label{Eq:PosteriorCircle2}%
\end{equation}
where the $s$ parameter is half the angular width of the overlap between the
posterior probabilities of adjacent neurons on the unit circle, in which case
$0\leq s\leq\frac{\pi}{M}$ ensures that no more than two neurons can respond
to a given input. Anticipating the optimum solution, a typical example of this
type of posterior probability is shown in figure \ref{Fig:PosteriorCircle2}.

In order to guarantee that $\Pr\left(  y|\mathbf{x}\right)  $ has a piecewise
linear dependence on $\mathbf{x}$, as is required of solutions of equation
\ref{Eq:StationaryP}, $f\left(  \theta\right)  $ must have the sinusoidal
dependence $f\left(  \theta\right)  =a+b\cos\theta+c\sin\left|  \theta\right|
$, where the use of $\left|  \theta\right|  $ arises because $p\left(
\theta\right)  =p\left(  -\theta\right)  $. Note that the $\Pr\left(
\mathbf{x}\right)  =0$ solution to the stationarity condition on $\Pr\left(
y|\mathbf{x}\right)  $ (see equation \ref{Eq:StationaryP}) implies that
$\Pr\left(  y|\mathbf{x}\right)  $ is undefined for any $\mathbf{x}$ that does
not lie on the unit circle. However, for those $\mathbf{x}$ that do lie on the
unit circle, the $a$, $b$ and $c$ parameters can be determined by demanding
continuity of $p\left(  \theta\right)  $ at the ends of its piecewise
intervals (i.e. at $\theta=\frac{\pi}{M}-s$ and $\theta=\frac{\pi}{M}+s$), and
by demanding that the total probability of any neuron firing first is unity
(i.e. the total posterior probability is normalised such that $f\left(
\theta\right)  +f\left(  \frac{2\pi}{M}-\theta\right)  =1$ in the interval
$\frac{\pi}{M}-s\leq\theta\leq\frac{\pi}{M}+s$), to obtain
\begin{equation}
f\left(  \theta\right)  =\frac{1}{2}+\frac{1}{2}\frac{\sin\left(  \frac{\pi
}{M}-\theta\right)  }{\sin s}\label{Eq:PosteriorCircle2Piece}%
\end{equation}
This corresponds to a piecewise linear contribution to $\Pr\left(
y|\mathbf{x}\right)  $ whose gradient points in the $\left(  -\sin\left(
\frac{\pi}{M}\right)  ,\cos\left(  \frac{\pi}{M}\right)  \right)  $ direction.
A typical example of this type of posterior probability is shown in figure
\ref{Fig:PosteriorCircle2}.

Without loss of generality (because the solution is symmetric under rotations
of $\theta$ which are multiples of $\frac{2\pi}{M}$) set $y=0$ in equation
\ref{Eq:StationaryP}, to obtain in the interval $\frac{\pi}{M}-s\leq\theta
\leq\frac{\pi}{M}+s$%
\begin{align}
0   =&r\csc^{2}s\,\,\sin\left(  \frac{\pi}{M}\right)  \sin\left(  \frac{\pi
}{M}-\theta\right)  \left(  \sin s-\sin\left(  \frac{\pi}{M}-\theta\right)
\right) \nonumber\\
& \times\left(  n\sin s-\left(  n-1\right)  \,r\sin\left(  \frac{\pi}%
{M}\right)  \right)  \,\,
\end{align}
which may be solved for the optimum length $r$ of the reference vectors, to
yield
\begin{equation}
r=\frac{n}{n-1}\frac{\sin s}{\sin\left(  \frac{\pi}{M}\right)  }%
\label{Eq:StationaryRCircle2}%
\end{equation}
Set $y=0$ in equation \ref{Eq:StationaryX2} to obtain a transcendental
equation that must be satisfied by the optimum $s$
\begin{equation}
\frac{\sin s}{\sin\left(  \frac{\pi}{M}\right)  }-\frac{n-1}{n}\frac{M}{\pi
}\sin\left(  \frac{\pi}{M}\right)  \left(  \cos s+s\sin s\right)
=0\label{Eq:StationarySCircle2}%
\end{equation}%

\sloppypar
The symmetry of the solution may be used to make the replacement $\frac
{1}{2\pi}\int_{0}^{2\pi}d\theta\,\left(  \cdots\right)  \rightarrow\frac
{M}{\pi}\int_{0}^{\frac{\pi}{M}}d\theta\,\left(  \cdots\right)  $ in the
expressions for $D_{1}$ and $D_{2}$, which may then be evaluated and
simplified to yield the minimum $D_{1}+D_{2}$ as
\begin{equation}
D_{1}+D_{2}=2-\frac{n}{n-1}\frac{M}{2\pi}\left(  2s+\sin\left(  2s\right)
\right) \label{Eq:StationaryD1D2Circle2}%
\end{equation}
The value of $s$ which should be used in this expression for $D_{1}+D_{2}$ is
the solution of equation \ref{Eq:StationarySCircle2} for the chosen values of
$M$ and $n$.

Note that the expression for $r$ in equation \ref{Eq:StationaryRCircle2} and
the expression for $D_{1}+D_{2}$ in equation \ref{Eq:StationaryD1D2Circle2}
both have a finite limits as $n\rightarrow1$, because the limiting behaviour
of the solution $s$ of equation \ref{Eq:StationarySCircle2} is $s\rightarrow
\left(  n-1\right)  \frac{M}{\pi}\sin^{2}\left(  \frac{\pi}{M}\right)  $ (see
the asymptotic results in section \ref{Sect:Asymptotic}), which contains a
factor $n-1$ to cancel the $\frac{1}{n-1}$ factor that appears in both
equation \ref{Eq:StationaryRCircle2} and equation
\ref{Eq:StationaryD1D2Circle2}.

\subsection{Three Overlapping Posterior Probabilities}

\label{Sect:CircularManifold3}
\begin{figure}
[h]
\begin{center}
\includegraphics{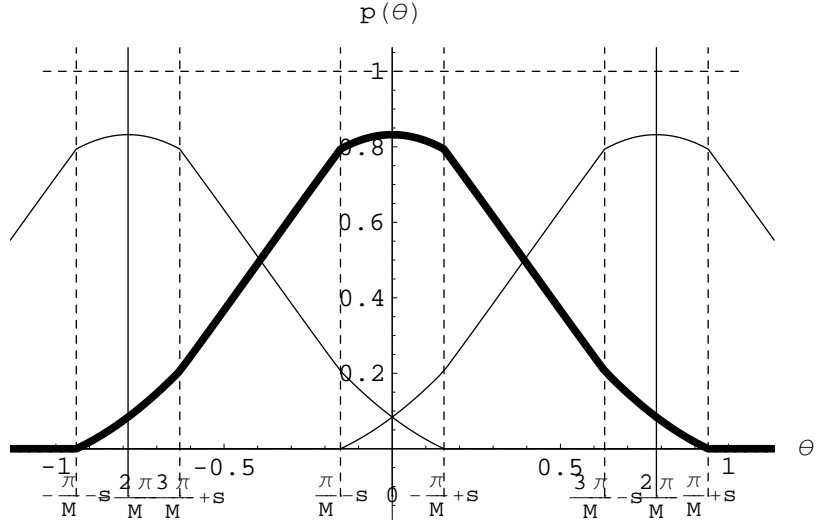}
\caption{\textit{Plot of the optimal neural posterior probability }$p\left( \theta\right)  $\textit{\ for
}$M=8$\textit{\ and }$n=100$\textit{. The
neighbouring posterior probabilities }$p\left(  \theta\pm\frac{2\pi}%
{M}\right)  $\textit{\ are also plotted. The optimal value of }$s
$\textit{\ is }$s\approx1.39\frac{\pi}{M}$\textit{.}}%
\label{Fig:PosteriorCircle3}%
\end{center}
\end{figure}

A detailed derivation of the results reported in
this section is given in appendix \ref{Appendix:CircularManifold3}. Because
the neurons have an angular separation of $\frac{2\pi}{M}$, the functional
form of $p\left(  \theta\right)  $ may be defined as%

\begin{equation}
p\left(  \theta\right)  =\left\{
\begin{array}
[c]{ll}%
f_{1}\left(  \theta\right)  & 0\leq\left|  \theta\right|  \leq-\frac{\pi}%
{M}+s\\
f_{2}\left(  \theta\right)  & -\frac{\pi}{M}+s\leq\left|  \theta\right|
\leq\frac{3\pi}{M}-s\\
f_{3}\left(  \theta\right)  & \frac{3\pi}{M}-s\leq\left|  \theta\right|
\leq\frac{\pi}{M}+s\\
0 & \left|  \theta\right|  \geq\frac{\pi}{M}+s
\end{array}
\right.
\end{equation}
where the $s$ parameter is half the angular width of the overlap between the
posterior probabilities of adjacent neurons on the unit circle, in which case
$\frac{\pi}{M}\leq s\leq\frac{2\pi}{M}$ ensures that no more than 3 neurons
can respond to a given input. Anticipating the optimum solution, a typical
example of this type of posterior probability is shown in figure
\ref{Fig:PosteriorCircle3}.

In order to guarantee that $\Pr\left(  y|\mathbf{x}\right)  $ has a piecewise
linear dependence on $\mathbf{x}$, the $f_{i}\left(  \theta\right)  $ must
have the sinusoidal dependence $f_{i}\left(  \theta\right)  =a_{i}+b_{i}%
\cos\theta+c_{i}\sin\left|  \theta\right|  $ for $i=1,2,3$. For those
$\mathbf{x}$ that lie on the unit circle, the $a_{i}$, $b_{i}$ and $c_{i}$
parameters can be determined by imposing continuity of $p\left(
\theta\right)  $ at $\theta=-\frac{\pi}{M}+s$, $\theta=\frac{3\pi}{M}-s$ and
$\theta=\frac{\pi}{M}+s$, and normalisation of the total posterior probability
such that $f_{1}\left(  \theta\right)  +f_{3}\left(  \frac{2\pi}{M}%
+\theta\right)  +f_{3}\left(  \frac{2\pi}{M}-\theta\right)  =1$ in the
interval $0\leq\theta\leq-\frac{\pi}{M}+s$, and $f_{2}\left(  \theta\right)
+f_{2}\left(  \frac{2\pi}{M}-\theta\right)  =1$ in the interval $-\frac{\pi
}{M}+s\leq\theta\leq\frac{3\pi}{M}-s$. Also, to satisfy the stationarity
conditions, set $y=0$ in equation \ref{Eq:StationaryX2}, and also set $y=0$ in
equation \ref{Eq:StationaryP} in each of the intervals $0\leq\theta\leq
-\frac{\pi}{M}+s$, $-\frac{\pi}{M}+s\leq\theta\leq\frac{3\pi}{M}-s$ and
$\frac{3\pi}{M}-s\leq\theta\leq\frac{\pi}{M}+s$. These conditions are
sufficient to solve for the optimum The $f_{i}\left(  \theta\right)  $ for
$i=1,2,3$, the optimum $r$, and the optimum $s$.

The optimum $f_{i}\left(  \theta\right)  $ are
\begin{align}
f_{1}\left(  \theta\right)   & =-\frac{1}{4}\left(  \cos\left(  \frac{4\pi}%
{M}-s\right)  +\cos s-2\cos\left(  \frac{\pi}{M}\right)  \cos\theta\right)
\csc^{2}\left(  \frac{\pi}{M}\right)  \sec\left(  \frac{2\pi}{M}-s\right)
\nonumber\\
f_{2}\left(  \theta\right)   & =\frac{1}{2}\left(  \cot\left(  \frac{\pi}%
{M}\right)  \sec\left(  \frac{2\pi}{M}-s\right)  \sin\left(  \frac{\pi}%
{M}-\theta\right)  +1\right) \nonumber\\
f_{3}\left(  \theta\right)   & =-\frac{1}{4}\csc^{2}\left(  \frac{\pi}%
{M}\right)  \left(  \cos\left(  \frac{3\pi}{M}-\theta\right)  \sec\left(
\frac{2\pi}{M}-s\right)  -1\right)
\end{align}
which correspond to different piecewise linear contributions to $\Pr\left(
y|\mathbf{x}\right)  $. The $f_{1}\left(  \theta\right)  $ piece has a
gradient that points in the $\left(  1,0\right)  $ direction, the
$f_{2}\left(  \theta\right)  $ piece has a gradient that points in the
$\left(  -\sin\left(  \frac{\pi}{M}\right)  ,\cos\left(  \frac{\pi}{M}\right)
\right)  $ direction, and the $f_{3}\left(  \theta\right)  $ piece has a
gradient that points in the $\left(  -\sin\left(  \frac{3\pi}{M}\right)
,\cos\left(  \frac{3\pi}{M}\right)  \right)  $ direction. The optimum $r$ is
\begin{equation}
r=\frac{n}{n-1}\frac{\cos\left(  \frac{2\pi}{M}-s\right)  }{\cos\left(
\frac{\pi}{M}\right)  }\label{Eq:StationaryRCircle3}%
\end{equation}
and the transcendental equation that must be satisfied by the optimum $s$ (for
$M=4$ this reduces to equation \ref{Eq:StationarySCircle2}) is
\begin{equation}
\frac{1}{n}\frac{\cos\left(  \frac{2\pi}{M}-s\right)  }{\cos\left(  \frac{\pi
}{M}\right)  }-\frac{n-1}{n}\frac{M}{\pi}\cos\left(  \frac{\pi}{M}\right)
\left(  \sin\left(  \frac{2\pi}{M}-s\right)  -\left(  \frac{2\pi}{M}-s\right)
\cos\left(  \frac{2\pi}{M}-s\right)  \right)  =0\label{Eq:StationarySCircle3}%
\end{equation}
and the minimum $D_{1}+D_{2}$ may be obtained as
\begin{align}
D_{1}+D_{2}   =&\frac{n\,\left(  \left(  n-1\right)  \left(  2\frac{n-2}%
{n}-\frac{M}{\pi}\,s\right)  -\sec^{2}\left(  \frac{\pi}{M}\right)  \right)
}{2\left(  n-1\right)  ^{2}}\nonumber\\
& -\frac{n\,\left(  \left(  n-1\right)  \left(  2-\frac{M}{\pi}\,s\right)
+\sec^{2}\left(  \frac{\pi}{M}\right)  \right)  }{2\left(  n-1\right)  ^{2}%
}\cos\left(  \frac{4\pi}{M}-2s\right) \label{Eq:StationaryD1D2Circle3}%
\end{align}

As in section \ref{Sect:CircularManifold2}, the limit $n\rightarrow1$ is well
behaved because the limiting behaviour of the solution $s$ of equation
\ref{Eq:StationarySCircle3} contains a factor $n-1$ (see the asymptotic
results in section \ref{Sect:Asymptotic}) to cancel the $\frac{1}{n-1}$ factor
that appears in both equation \ref{Eq:StationaryRCircle3} and equation
\ref{Eq:StationaryD1D2Circle3}.%

\begin{figure}
[h]
\begin{center}
\includegraphics{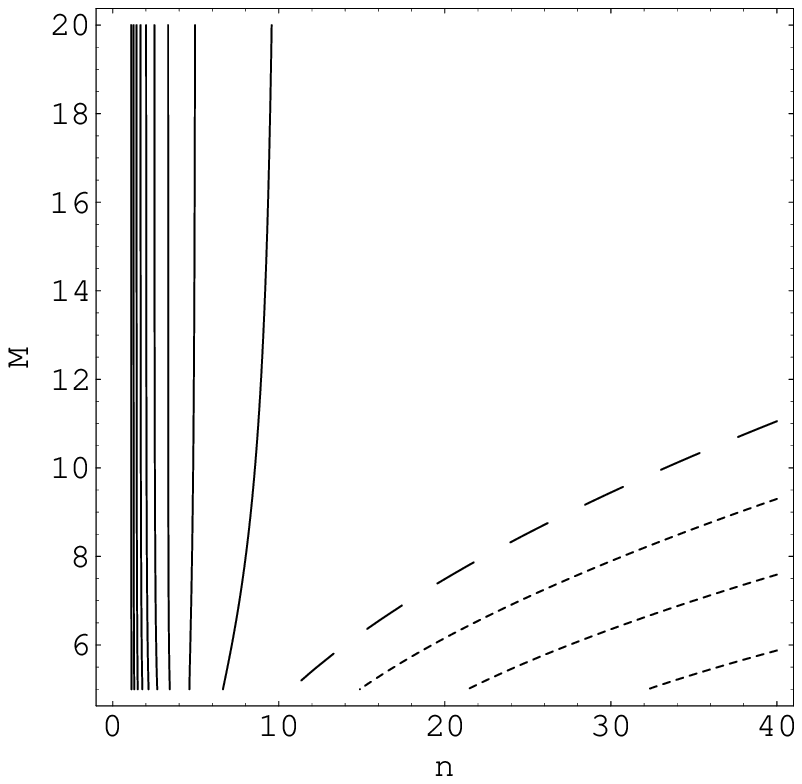}
\caption{\textit{Contour plot of the optimum value of }$s$\textit{\ versus }$\left(  n,M\right)  $\textit{\ for
encoding of a circular manifold. The solid contours are for the interval }$0\leq s<\frac{\pi}{M}$\textit{, the
dotted contours are for }$\frac{\pi}{M}<s\leq\frac{2\pi}{M}$\textit{, and the dashed contour is for
}$s=\frac{\pi}{M}$\textit{\ (this behaves asymptotically as }$n\approx3\frac{M^{2}}{\pi^{2}}$\textit{). The
contours are all separated
by intervals of }$\frac{\pi}{10M}$\textit{.}}%
\label{Fig:OverlapCircle}%
\end{center}
\end{figure}

The results for the optimum value of $s$ (i.e. equation
\ref{Eq:StationarySCircle2} and equation \ref{Eq:StationarySCircle3})\ may be
combined to yield the results shown in figure \ref{Fig:OverlapCircle}.

Asymptotically, as $M\rightarrow\infty$ and $n\rightarrow\infty$, the contour
$s=\frac{\pi}{M}$ (the dashed line in figure \ref{Fig:OverlapCircle}), which
is the boundary between the regions where 2 and 3 posterior probabilities
overlap, is given by $n\approx3\frac{M^{2}}{\pi^{2}}$ (see the asymptotic
results in section \ref{Sect:Asymptotic}).

\begin{figure}
[h]
\begin{center}
\includegraphics{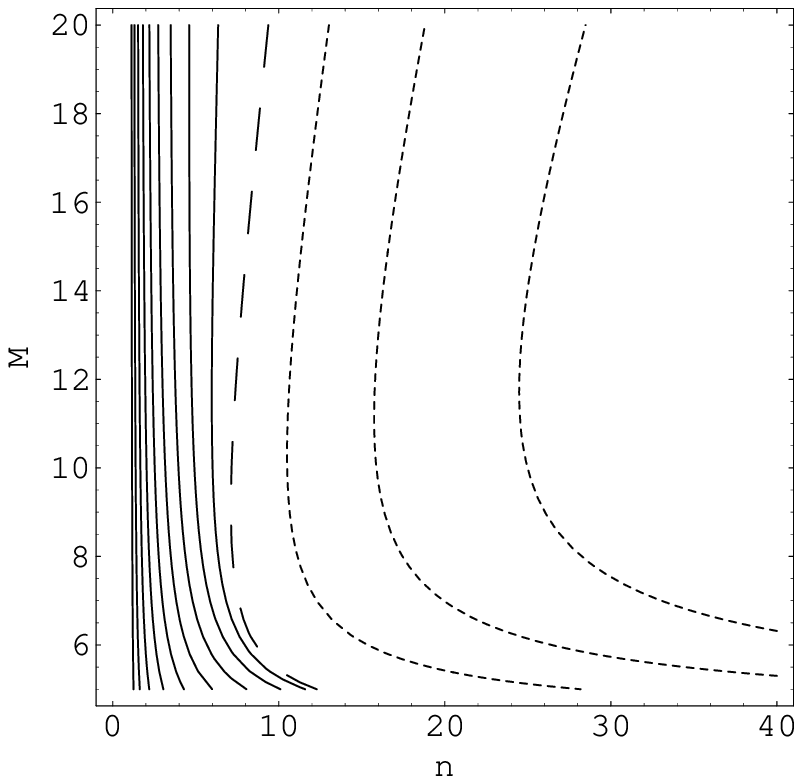}
\caption{\textit{Contour plot of the optimum value of }$s$\textit{\ versus }$\left(  n,M\right)  $\textit{\ for
joint encoding of a toroidal manifold.
The solid contours are for the interval }$0\leq s<\frac{\pi}{\sqrt{M}}%
$\textit{, the dotted contours are for }$\frac{\pi}{\sqrt{M}}<s\leq\frac{2\pi
}{\sqrt{M}}$\textit{, and the dashed contour is for }$s=\frac{\pi}{\sqrt{M}}$
\textit{(this behaves asymptotically as }$n\approx3\frac{M}{\pi^{2}}%
$\textit{). The contours are all separated by intervals of }$\frac{\pi
}{10\sqrt{M}}$\textit{.}}%
\label{Fig:OverlapTorusJoint}%
\end{center}
\end{figure}

The corresponding results for joint encoding of input vectors that live on a
2-torus are shown in figure \ref{Fig:OverlapTorusJoint}.%

\section{Toroidal Manifold: Factorial Encoding}

\label{Sect:ToroidalManifold}All of the results for factorial encoding of
input data that lives on a toroidal manifold may be derived from the
expression for $D_{1}+D_{2}$ in equation \ref{Eq:D1D2Factorial} (and the
corresponding stationarity conditions), with the appropriate replacements for
equations \ref{Eq:UnitCircle}, \ref{Eq:PosteriorCircle} and \ref{Eq:RV}.

The posterior probability $p\left(  \theta\right)  $ then has the same
functional form as for a circular manifold, except that $M$ is replaced by
$\frac{M}{2}$ because each of the two dimensions uses exactly half of the
total of $M$ neurons, so these results are not quoted explicitly here. The
steps in the derivation of the optimum values of $r$ and $s$ and the minimum
value of $D_{1}+D_{2}$ are analogous to the steps that appear in the
derivation for a circular input manifold, and the results are sufficiently
different from the ones that were obtained from a circular manifold that they
are quoted explicitly here.

\subsection{Two Overlapping Posterior Probabilities}

\label{Sect:ToroidalManifold2}A detailed derivation of the results reported in
this section is given in appendix \ref{Appendix:ToroidalManifold2}. The
stationarity conditions yield the optimum $r$ as
\begin{equation}
r=\frac{2n}{n-1}\frac{\sin s}{\sin\left(  \frac{2\pi}{M}\right)
}\label{Eq:StationaryRTorus2}%
\end{equation}
The transcendental equation that must be satisfied by the optimum $s$ is
\begin{equation}
\frac{\sin s}{\sin\left(  \frac{2\pi}{M}\right)  }-\frac{n-1}{n+1}\frac
{M}{2\pi}\sin\left(  \frac{2\pi}{M}\right)  \ \left(  \cos s+s\sin s\right)
=0\label{Eq:StationarySTorus2}%
\end{equation}
The expression for the minimum $D_{1}+D_{2}$ is
\begin{equation}
D_{1}+D_{2}=4-\frac{n}{n-1}\frac{M}{2\pi}\left(  2s+\sin\left(  2s\right)
\right) \label{Eq:StationaryD1D2Torus2}%
\end{equation}

\subsection{Three Overlapping Posterior Probabilities}

\label{Sect:ToroidalManifold3}A detailed derivation of the results reported in
this section is given in appendix \ref{Appendix:ToroidalManifold3}. The
stationarity conditions yield the optimum $r$ as
\begin{equation}
r=\frac{2n}{n-1}\frac{\cos\left(  \frac{4\pi}{M}-s\right)  }{\cos\left(
\frac{2\pi}{M}\right)  }\ \label{Eq:StationaryRTorus3}%
\end{equation}
The transcendental equation that must be satisfied by the optimum $s$ is
\begin{equation}
\frac{1}{n}\frac{\cos\left(  \frac{4\pi}{M}-s\right)  }{\cos\left(  \frac
{2\pi}{M}\right)  }-\frac{n-1}{2n}\frac{M}{2\pi}\cos\left(  \frac{2\pi}%
{M}\right)  \left(  \sin\left(  \frac{4\pi}{M}-s\right)  -\left(  \frac{4\pi
}{M}-s\right)  \ \cos\left(  \frac{4\pi}{M}-s\right)  \right)
=0\label{Eq:StationarySTorus3}%
\end{equation}
The expression for the minimum $D_{1}+D_{2}$ is
\begin{align}
D_{1}+D_{2}   =&\frac{n\left(  \left(  n-1\right)  \left(  2\frac{n-2}%
{n}\ -\frac{M}{2\pi}\,s\right)  -2\sec^{2}\left(  \frac{2\pi}{M}\right)
\right)  }{\left(  n-1\right)  ^{2}}\nonumber\\
& -\frac{n\left(  \left(  n-1\right)  \ \left(  2-\frac{M}{2\pi}\,s\right)
+2\sec^{2}\left(  \frac{2\pi}{M}\right)  \right)  }{\left(  n-1\right)  ^{2}%
}\cos\left(  \frac{8\pi}{M}-2s\right) \label{Eq:StationaryD1D2Torus3}%
\end{align}

\begin{figure}
[h]
\begin{center}
\includegraphics{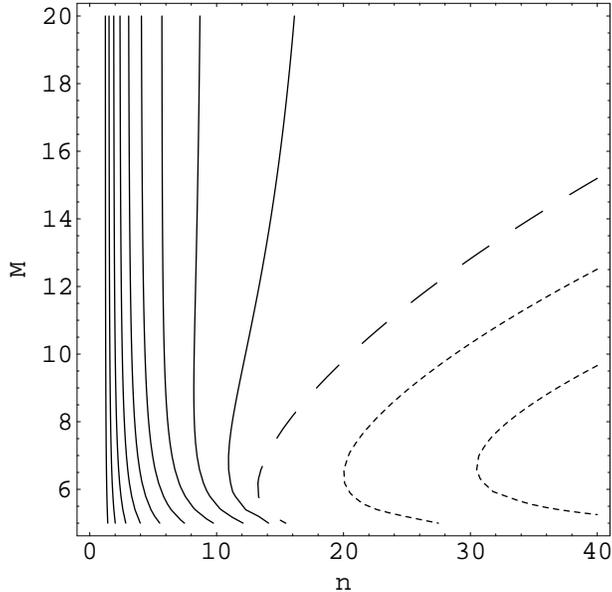}
\caption{\textit{Contour plot of the optimum value of }$s$\textit{\ versus }$\left(  n,M\right)  $\textit{\ for
factorial encoding of a toroidal
manifold. The solid contours are for the interval }$0\leq s<\frac{2\pi}{M}%
$\textit{, the dotted contours are for }$\frac{2\pi}{M}<s\leq\frac{4\pi}{M}%
$\textit{, and the dashed contour is for }$s=\frac{2\pi}{M}$\textit{\ (this
behaves asymptotically as }$n\approx\frac{3}{2}\frac{M^{2}}{\pi^{2}}%
$\textit{). The contours are all separated by intervals of }$\frac{\pi}{5M}%
$\textit{.}}%
\label{Fig:OverlapTorus}%
\end{center}
\end{figure}

The results for the optimum value of $s$ (i.e. equation
\ref{Eq:StationarySTorus2} and equation \ref{Eq:StationarySTorus3})\ may be
combined to yield the results shown in figure \ref{Fig:OverlapTorus}.%

\section{Joint Versus Factorial Encoding}

\label{Sect:JointVsFactorial}
\begin{figure}
[h]
\begin{center}
\includegraphics{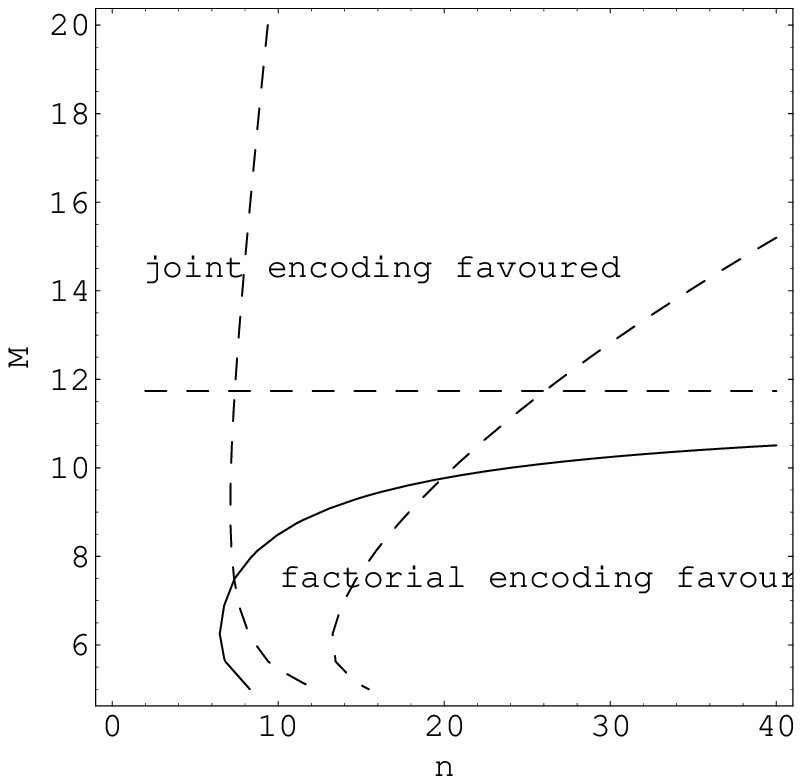}
\caption{\textit{The diagram shows various results pertaining to joint and factorial encoding of a 2-torus. The
solid line is the boundary between the regions of the }$\left(  n,M\right)  $\textit{\ plane where joint or
factorial encoding are favoured, and the horizontal dashed line is the asymptotic limit }$M\approx12$\textit{\
of this boundary as }$n\rightarrow\infty$\textit{. The left hand dashed line is the boundary between the regions
where 2 or 3 overlapping posterior probabilites occur in joint encoding, and the right hand
dashed line is the corresponding boundary for factorial encoding.}}%
\label{Fig:Stability}%
\end{center}
\end{figure}

The results in section
\ref{Sect:CircularManifold} and section \ref{Sect:ToroidalManifold} may be
used to deduce when a factorial encoder is favoured with respect to a joint
encoder (for input data that lives on a 2-torus). Firstly, equation
\ref{Eq:StationarySCircle2} (with the replacement $M\rightarrow\sqrt{M}$, and
setting $s=\frac{\pi}{\sqrt{M}}$)\ may be used to deduce the region of the
$\left(  n,M\right)  $ plane where joint encoding of a 2-torus involves no
more that $2$ overlapping posterior probabilities, and equation
\ref{Eq:StationarySTorus2} (with $s=\frac{2\pi}{M}$) may be used to deduce the
corresponding result for factorial encoding of a 2-torus. Once these regions
have been established, it is then possible to decide which of equation
\ref{Eq:StationaryD1D2Circle2} or equation \ref{Eq:StationaryD1D2Circle3}
(with $M\rightarrow\sqrt{M}$ and then multiplied overall by $2$) to use to
calculate $D_{1}+D_{2}$ in the case of joint encoding a 2-torus, and which of
equation \ref{Eq:StationaryD1D2Torus2} or equation
\ref{Eq:StationaryD1D2Torus3} to use to calculate $D_{1}+D_{2}$ in the case of
factorial encoding a 2-torus. These results are gathered together in figure
\ref{Fig:Stability}.%

\begin{figure}
[h]
\begin{center}
\includegraphics{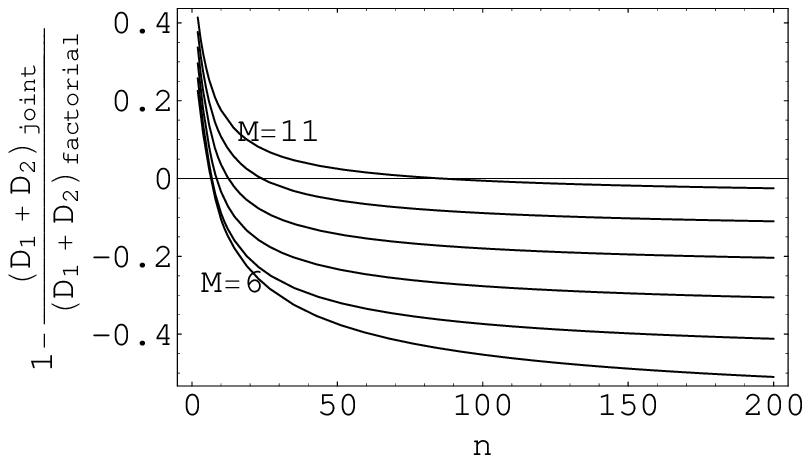}
\caption{\textit{Plots for }$M=6,7,8,9,10,11$\textit{\ of }$\left(
D_{1}+D_{2}\right)  _{factorial}-\left(  D_{1}+D_{2}\right)  _{joint}%
$\textit{\ in units in which }$\left(  D_{1}+D_{2}\right)  _{factorial}%
=1$\textit{. This makes it clear that the degree to which a factorial encoder
is favoured with respect to a joint encoder is quite significant for large
}$n$\textit{.}}%
\label{Fig:D1D2JointFactorial}%
\end{center}
\end{figure}

The need to derive results where up to 3 posterior probabilities overlap
(which involves a large amount of algebra) is clear from the results shown in
figure \ref{Fig:Stability}, where it may be seen that most of the region where
the factorial encoder is favoured with respect to the joint encoder has up to
3 overlapping posterior probabilities. The degree to which a factorial encoder
is favoured with respect to a joint encoder may be seen in figure
\ref{Fig:D1D2JointFactorial}.

If the number of neurons $M$ is restricted (i.e. $M\lesssim12$), then the
joint encoding scheme in which the 2-torus is encoded using small encoding
cells as shown in figure \ref{Fig:EncodeCurved}(a), is usually not as good as
the factorial encoding scheme in which the 2-torus is encoded using the
intersection of pairs of elongated encoding cells as shown in figure
\ref{Fig:EncodeCurved}(b). This does require that the number of firing events
$n$ is sufficiently large that both subsets of $\frac{M}{2}$ neurons in the
factorial encoder are virtually guaranteed to each receive at least 1 firing
event, so that they can indeed approximate the input vector by the
intersection of a pair of response regions.

If the number of neurons $M$ is too large (i.e. $M\gtrsim12$), then the joint
encoding scheme is always favoured with respect to the factorial encoding
scheme, because there are sufficient neurons to encode the 2-torus well using
small response regions, as shown in figure \ref{Fig:EncodeCurved}(a). This
includes the limiting case $M\rightarrow\infty$, where the curvature of the
input manifold is not visible to each neuron separately, because each neuron
then responds to an infinitesimally small angular interval of the input
manifold. This result implies that joint encoding is always favoured when the
input manifold is planar, as was discussed in figure \ref{Fig:EncodeLinear}
and figure \ref{Fig:EncodeCurved}.

Although not presented here, these results generalise readily to higher
dimensional toruses, where factorial encoding is even more favoured, because
(roughly speaking) the number of neurons required to do joint encoding with a
given resolution\ increases exponentially with the dimensionality of the
input, whereas the number of neurons required to do factorial encoding with a
given resolution\ increases linearly with the dimensionality of the input
(provided that enough firing events are observed).

\section{Asymptotic Results}

\label{Sect:Asymptotic}Referring to figure \ref{Fig:Stability}, the asymptotic
behaviour as $M\rightarrow\infty$ lies in the region where two posterior
probabilities overlap, and the asymptotic behaviour as $n\rightarrow\infty$
lies in the region where three posterior probabilities overlap, so care must
be taken to use the appropriate results when deriving the various asymptotic
approximations below. The boundary between the regions where two or three
posterior probabilities overlap can be obtained for a circular input manifold
by putting $s=\frac{\pi}{M}$ in equation \ref{Eq:StationarySCircle2} (or
$s=\frac{2\pi}{M}$ in equation \ref{Eq:StationarySTorus2} in the case of a
toroidal input manifold), and as $M\rightarrow\infty$ this is given by
\begin{equation}
n\approx\left\{
\begin{array}
[c]{ll}%
3\frac{M^{2}}{\pi^{2}} & \text{circular manifold}\\
\frac{3}{2}\frac{M^{2}}{\pi^{2}} & \text{toroidal manifold (factorial
encoding)}%
\end{array}
\right.
\end{equation}

As $M\rightarrow\infty$ the asymptotic behaviour of $D_{1}+D_{2}$ for a
circular input manifold may be obtained by asymptotically expanding the $s$
dependence of equation \ref{Eq:StationarySCircle2} (or equation
\ref{Eq:StationarySTorus2} in the case of a toroidal input manifold) in
inverse powers of $M$, to yield%

\begin{equation}
s\approx\left\{
\begin{array}
[c]{ll}%
\frac{n-1}{n}\frac{\pi}{M}+\frac{(n-1)(n^{2}-4n+2)}{3n^{3}}\frac{\pi^{3}%
}{M^{3}} & \text{circular manifold}\\
\frac{n-1}{n+1}\frac{2\ \pi}{M}+\frac{\ (n-1)(n^{2}-6n+1)}{3(n+1)^{3}}\left(
\frac{2\ \pi}{M}\right)  ^{3} & \text{toroidal manifold (factorial encoding)}%
\end{array}
\right.
\end{equation}
and substituting this solution into the appropriate expression for $r$ to
obtain
\begin{equation}
r\approx\left\{
\begin{array}
[c]{ll}%
1+\frac{(2n^{2}-6n+3)}{6n^{2}}\frac{\pi^{2}}{M^{2}} & \text{circular
manifold}\\
\bigskip\frac{2n}{n+1}+\frac{8n(n^{2}-4n+1)}{3(n+1)^{3}}\frac{\pi^{2}}{M^{2}}%
& \text{toroidal manifold (factorial encoding)}%
\end{array}
\right.
\end{equation}
and substituting this solution into the appropriate expression for
$D_{1}+D_{2}$ to obtain
\begin{equation}
D_{1}+D_{2}\approx\left\{
\begin{array}
[c]{ll}%
\frac{2(2n-1)\ }{3n^{2}}\frac{\pi^{2}}{M^{2}} & \text{circular manifold}\\
\frac{4}{n+1}+\frac{64n^{2}}{3(n+1)^{3}}\frac{\ \pi^{2}}{M^{2}} &
\text{toroidal manifold (factorial encoding)}%
\end{array}
\right.
\end{equation}
The asymptotic result for a circular manifold may be used to determine the
corresponding result for a linear manifold. Thus, if lengths are scaled so
that the separation of the neurons (as measured around the circular manifold)
becomes unity, which requires that all lengths are divided by $\frac{2\ \pi
}{M}$, then asymptotically as $M\rightarrow\infty$ the circular manifold
solution becomes identical to the solution for a linear manifold with neurons
separated by unit distance. Thus the optimum solution for a linear manifold
with neurons separated by unit distance is $s=\frac{n-1}{2n}$ and $D_{1}%
+D_{2}=\frac{2n-1\ }{6n^{2}}$ (note that $D_{1}+D_{2}$ has the dimensions of
$\left(  \text{length}\right)  ^{2}$).

As $n\rightarrow1$ (i.e. the LBG\ vector quantiser limit)\ the asymptotic
behaviour of $D_{1}+D_{2}$ for a circular input manifold may be obtained by
expanding the $s$ dependence of equation \ref{Eq:StationarySCircle2} about the
point $s=0$ (or equation \ref{Eq:StationarySTorus2} about the point $s=0$ for
a toroidal input manifold), to yield
\begin{equation}
s\approx\left\{
\begin{array}
[c]{ll}%
\left(  n-1\right)  \frac{M}{\pi}\sin^{2}\left(  \frac{\pi}{M}\right)  &
\text{circular manifold}\\
\frac{n-1}{2}\frac{M}{2\pi}\sin^{2}\left(  \frac{2\pi}{M}\right)  &
\text{toroidal manifold (factorial encoding)}%
\end{array}
\right.
\end{equation}
which gives $s=0$ when $n=1$, so there is no overlap between the posterior
probabilities for different neurons, as would be expected in a vector
quantiser where only one neuron is allowed to fire. Substitute this solution
into the appropriate expression for $r$ to obtain at $n=1$%
\begin{equation}
r=\left\{
\begin{array}
[c]{ll}%
\frac{M}{\pi}\sin\left(  \frac{\pi}{M}\right)  & \text{circular manifold}\\
\frac{M\ }{2\pi}\sin\left(  \frac{2\pi}{M}\right)  & \text{toroidal manifold
(factorial encoding)}%
\end{array}
\right.
\end{equation}
which is the distance of the centroid of an arc of the unit circle (with
angular length $\frac{2\pi}{M}$ for a circular manifold, or angular length
$\frac{4\pi}{M}$ for a toroidal manifold) from the origin, as expected for a
network in which only one neuron can fire. So the best reconstruction is the
centroid of the inputs that could have caused the single firing event. These
results may be substituted into the appropriate expression for $D_{1}+D_{2}$
to obtain at $n=1$
\begin{equation}
D_{1}+D_{2}=\left\{
\begin{array}
[c]{ll}%
2-2\left(  \frac{M}{\pi}\right)  ^{2}\sin^{2}\left(  \frac{\pi}{M}\right)  &
\text{circular manifold}\\
4-2\left(  \frac{M}{2\pi}\right)  ^{2}\sin^{2}\left(  \frac{2\pi}{M}\right)  &
\text{toroidal manifold (factorial encoding)}%
\end{array}
\right.
\end{equation}
These results for $D_{1}+D_{2}$ have a simple geometrical interpretation. For
a circular manifold $D_{1}+D_{2}$ is (twice) the average squared distance from
an arc with angular length $\frac{2\pi}{M}$ to its associated reference
vector, which is exactly what would be expected. For a toroidal manifold
$D_{1}+D_{2}$ is the same result with $M\rightarrow\frac{M}{2}$, plus an extra
contribution of 2, because a factorial encoder with only 1 firing event acts
as a conventional encoder using $\frac{M}{2}$ neurons for the circular
dimension that is fortunate enough to be associated with the firing event
(hence the first contribution to $D_{1}+D_{2}$), and acts as no encoder at all
for the other circular dimension which is associated with no firing events
(hence the extra contribution of 2 to $D_{1}+D_{2}$).

As $n\rightarrow\infty$ the asymptotic behaviour of $D_{1}+D_{2}$ for a
circular input manifold may be obtained by expanding the $s$ dependence of
equation \ref{Eq:StationarySCircle3} about the point $s=\frac{2\pi}{M}$ (or
equation \ref{Eq:StationarySTorus3} about the point $s=\frac{4\pi}{M}$ for a
toroidal input manifold), to yield
\begin{equation}
s\approx\left\{
\begin{array}
[c]{ll}%
\frac{2\pi}{M}-\left(  \frac{3\pi}{M\,n\cos^{2}\left(  \frac{\pi}{M}\right)
}\right)  ^{\frac{1}{3}} & \text{circular manifold}\\
\frac{4\pi}{M}-\left(  \frac{12\pi}{M\,n\cos^{2}\left(  \frac{2\pi}{M}\right)
}\right)  ^{\frac{1}{3}} & \text{toroidal manifold (factorial encoding)}%
\end{array}
\right.
\end{equation}
where the limiting values of $s$ as $n\rightarrow\infty$ (i.e. $s\rightarrow
\frac{2\pi}{M}$ for a circular manifold, and $s\rightarrow\frac{4\pi}{M}$ for
a toroidal manifold) stops just short of allowing four or more posterior
probabilities to overlap. In this limit $D_{1}=0$, so for a circular manifold
the network acts as a PCA encoder (see the discussion after equation
\ref{Eq:ObjectiveD1D2}) whose expansion coefficients sum to unity. In order to
encode vectors on a unit circle without error three basis vectors are
required; the expansion coefficients are probabilities which must sum to
unity, so three basis vectors are required in order that there are two
independent expansion coefficients. This is the reason why it is sufficient to
consider no more than three overlapping posterior probabilities for encoding
data that lives in a 2-dimensional manifold (this argument generalises
straightforwardly to higher dimensions). The same argument applies to the case
of factorial encoding of a toroidal manifold. Substitute this solution into
the appropriate expression for $r$ to obtain
\begin{equation}
r\approx \left\{ \begin{array}{cc}
 \frac{1}{2} \sec ( \frac{\pi }{\mathit{M}})  \left( 2-{\left( \frac{3\pi
}{\mathit{M} \mathit{n} {\cos }^{2}( \frac{\pi }{\mathit{M}}) }\right)
}^{2/3}\right)  & \mathrm{circular} \;\mathrm{manifold} \\
 \sec ( \frac{2\pi }{\mathit{M}})  \left( 2-{\left( \frac{12\pi
}{\mathit{M} \mathit{n} {\cos }^{2}( \frac{2\pi }{\mathit{M}}) }\right) }^{2/3}\right)  & \mathrm{toroidal}\;
\mathrm{manifold} \left( \mathrm{factorial}\; \mathrm{encoding}\right)
\end{array}\right.
\end{equation}
and substitute these results into the appropriate expression for $D_{1}+D_{2}
$ to obtain
\begin{equation}
D_{1}+D_{2}\approx\left\{
\begin{array}
[c]{ll}%
\frac{2\ }{n}\tan^{2}\left(  \frac{\pi}{M}\right)  & \text{circular
manifold}\\
\frac{4}{n}\left(  2\sec^{2}\left(  \frac{2\ \pi}{M}\right)  -1\right)  &
\text{toroidal manifold (factorial encoding)}%
\end{array}
\right.
\end{equation}
Thus as $n\rightarrow\infty$ it is possible to derive a value of $M$ for which
the asymptotic $D_{1}+D_{2}$ is the same for joint and factorial encoding of a
toroidal manifold. This value of $M$ must satisfy $\frac{4\ }{n}\tan
^{2}\left(  \frac{\pi}{\sqrt{M}}\right)  =\frac{4}{n}\left(  2\sec^{2}\left(
\frac{2\ \pi}{M}\right)  -1\right)  $, which yields $M\approx11.74 $.

\section{Approximate the Posterior Probability}

\label{Sect:ISO}A posterior probability may always be written in the form
\begin{equation}
\Pr\left(  y|\mathbf{x}\right)  =\frac{Q\left(  \mathbf{x}|y\right)  }%
{\sum_{y^{\prime}=0}^{M-1}Q\left(  \mathbf{x}|y^{\prime}\right)  }%
\end{equation}
where $Q\left(  \mathbf{x}|y\right)  \geq0$ (with $Q\left(  \mathbf{x}%
|y\right)  >0$ for at least one value of $y$ for each $\mathbf{x}$). If the
neurons behaved in such a way that they produced independent Poissonian firing
events in response to a given input, then $Q\left(  \mathbf{x}|y\right)  $
would be the firing rate (or activation function) of neuron $y$ in response to
input $\mathbf{x}$.

The optimum solution $p\left(  \theta\right)  $ (as given in equation
\ref{Eq:PosteriorCircle2} and equation \ref{Eq:PosteriorCircle2Piece}) may be
approximated on the unit circle (i.e. $\mathbf{x}=\left(  \cos\theta
,\sin\theta\right)  $) by defining $Q\left(  \mathbf{x}|y\right)  $ as
\begin{align}
Q\left(  \mathbf{x}|y\right)   & =\left\{
\begin{array}
[c]{ll}%
\mathbf{w}\cdot\mathbf{x}-a & \mathbf{w}\cdot\mathbf{x}\geq a\\
0 & \mathbf{w}\cdot\mathbf{x}\leq a
\end{array}
\right. \nonumber\\
\mathbf{w}  & =\left(  \cos\left(  \frac{2\pi y}{M}\right)  ,\sin\left(
\frac{2\pi y}{M}\right)  \right) \nonumber\\
a  & =\cos\left(  \frac{\pi}{M}\right)  -\sin\left(  \frac{\pi}{M}\right)
\sin s\label{Eq:ActivationISO}%
\end{align}
where $a$ is a threshold parameter, and $\mathbf{w}$ is a unit weight vector.
This is the form of the neural activation function that is used in
\cite{Ref:ISO}. This leads to a good approximation to the optimum solution
$p\left(  \theta\right)  $ because
\begin{equation}
p\left(  \theta\right)  =\left\{
\begin{array}
[c]{ll}%
0 & \theta\leq-\frac{\pi}{M}-s\\
\frac{Q\left(  \mathbf{x}|y=0\right)  }{Q\left(  \mathbf{x}|y=0\right)
+Q\left(  \mathbf{x}|y=M-1\right)  }+O\left(  \left(  \theta+\frac{\pi}%
{M}\right)  ^{3}\right)  & -\frac{\pi}{M}-s\leq\theta\leq-\frac{\pi}{M}+s\\
1 & -\frac{\pi}{M}+s\leq\theta\leq\frac{\pi}{M}-s\\
\frac{Q\left(  \mathbf{x}|y=0\right)  }{Q\left(  \mathbf{x}|y=0\right)
+Q\left(  \mathbf{x}|y=1\right)  }+O\left(  \left(  \theta-\frac{\pi}%
{M}\right)  ^{3}\right)  & \frac{\pi}{M}-s\leq\theta\leq\frac{\pi}{M}+s\\
0 & \theta\geq\frac{\pi}{M}+s
\end{array}
\right.
\end{equation}
This approximation works well because curved input manifolds can be optimally
encoded by using appropriate hyperplanes (as defined in equation
\ref{Eq:ActivationISO})\ to slice off pieces of the manifold.

This approximation breaks down as $M\longrightarrow\infty$, as can be seen by
inspecting the series expansion of $p\left(  \theta\right)  $ near
$\theta=\frac{\pi}{M}$.
\begin{equation}
p\left(  \theta\right)  =\left\{
\begin{array}
[c]{ll}%
\begin{array}
[c]{l}%
\frac{1}{2}-\frac{1}{2}\frac{1}{\sin s}\left(  \theta-\frac{\pi}{M}\right)
+\frac{1}{12}\frac{1}{\sin s}\left(  \theta-\frac{\pi}{M}\right)  ^{3}\\
+O\left(  \left(  \theta-\frac{\pi}{M}\right)  ^{4}\right)
\end{array}
& \text{exact}\\%
\begin{array}
[c]{l}%
\frac{1}{2}-\frac{1}{2}\frac{1}{\sin s}\left(  \theta-\frac{\pi}{M}\right)
+\frac{1}{12}\left(  \frac{1}{\sin s}-\frac{3}{\tan\frac{\pi}{M}\sin^{2}%
s}\right)  \left(  \theta-\frac{\pi}{M}\right)  ^{3}\\
+O\left(  \left(  \theta-\frac{\pi}{M}\right)  ^{4}\right)
\end{array}
& \text{approximate}%
\end{array}
\right.
\end{equation}
which differ in the $O\left(  \left(  \theta-\frac{\pi}{M}\right)
^{3}\right)  $ term. In the limit $M\longrightarrow\infty$ the half-width
parameter $s$ behaves like $M^{-1\,}$, so the $O\left(  \left(  \theta
-\frac{\pi}{M}\right)  ^{3}\right)  $ term behaves like $M\left(  \theta
-\frac{\pi}{M}\right)  ^{3}$ in the exact case, and \thinspace$M^{3}\left(
\theta-\frac{\pi}{M}\right)  ^{3}$ in the approximate case because of the
contribution from the $\frac{3}{\tan\frac{\pi}{M}\sin^{2}s}$ term. As
$M\longrightarrow\infty$ each neuron responds to a progressively smaller
angular range of inputs on the unit circle, so from the point of view of each
neuron the curvature of the input manifold becomes negligible (i.e. the input
manifold appears to more and more closely approximate a straight line), which
ultimately makes it impossible to use hyperplanes to slice off pieces of the
manifold. In the $M\longrightarrow\infty$ limit, a better approximation to the
posterior probability would be to use ball-shaped regions (e.g. a radial basis
function network) to cut up the input manifold into pieces.

\section{Conclusions}

The results in this paper demonstrate that, for input data that lies on a
curved manifold (specifically, a 2-torus),\ and for an objective function that
measures the average reconstruction error (in the Euclidean sense)\ of a
2-layer neural network encoder, the type of encoder that is optimal depends on
the total number of neurons and on the total number of observed firing events
in the network output layer. There are two basic types of encoder: a joint
encoder in which the network acts as a vector quantiser for the whole input
space, and a factorial encoder in which the network breaks into a number of
subnetworks, each of which acts as a vector quantiser for a subspace of the
input space.

The particular conditions under which factorial encoding is favoured with
respect to joint encoding arise when the input data is derived from a curved
input manifold, provided that the number of neurons is not too large, and
provided that the number of observed neural firing events is large enough.
Factorial encoding does not emerge when the input manifold is insufficiently
curved, or equivalently when there are too many neurons, because then each
neuron does not have a sufficiently large encoding cell to be aware of the
manifold's curvature.

Factorial encoding allows the input data to be encoded using a much smaller
number of neurons than would be the case if joint encoding were used. Because
only a small number of neurons is used, a factorial encoding scheme must be
succinct, so it has to abstract the underlying degrees of freedom in the input
manifold; this is a very useful side-effect of factorial encoding. This effect
becomes stronger as the dimensionality of the curved input manifold is increased.

The main simplification that makes these calculations possible is that, in an
optimal neural network, the form for the posterior probability is a piecewise
linear function of the input vector. This leads to an enormous simplification
in the mathematics, because only the space of piecewise linear functions needs
to be searched for the optimal solution, rather than the whole space of
functions (subject to normalisation and non-negativity constraints).

A convenient approximation to this type of factorial encoder is the
partitioned mixture distribution (PMD) network \cite{Ref:PMD}, in which the
individual subnetworks in the factorial encoder network are constrained to
share parameters, which thus leads to an upper bound on the minimum value of
the objective function that would have ideally been obtained with the
unconstrained factorial encoder network.

\section{Acknowledgements}

I thank Chris Webber for many useful conversations that we had during the
course of this research.

\appendix

\section{Objective Function}

\label{Appendix:Objective}The objective function $D=2D_{VQ}$ is given by
\begin{equation}
D\equiv2\int d\mathbf{x}\Pr\left(  \mathbf{x}\right)  \sum_{\mathbf{y}}%
\Pr\left(  \mathbf{y}|\mathbf{x}\right)  \left\|  \mathbf{x}-\mathbf{x}%
^{\prime}\left(  \mathbf{y}\right)  \right\|  ^{2}%
\end{equation}
If the observed state of the output layer is the locations of $n$ firing
events on $M$ neurons, then this expression for $D$ can be manipulated into
the following form \cite{Ref:D1D2}
\begin{equation}
D=2\int d\mathbf{x}\Pr\left(  \mathbf{x}\right)  \sum_{y_{1}=1}^{M}\sum
_{y_{2}=1}^{M}\cdots\sum_{y_{n}=1}^{M}\Pr\left(  y_{1},y_{2},\cdots
,y_{n}|\mathbf{x}\right)  \left\|  \mathbf{x}-\mathbf{x}^{\prime}\left(
y_{1},y_{2},\cdots,y_{n}\right)  \right\|  ^{2}\label{Eq:Objective}%
\end{equation}
where $\Pr\left(  \mathbf{y}|\mathbf{x}\right)  $ has now been replaced by the
more explicit notation $\Pr\left(  y_{1},y_{2},\cdots,y_{n}|\mathbf{x}\right)
$, and $\mathbf{x}^{\prime}\left(  y_{1},y_{2},\cdots,y_{n}\right)  $ is a
vector given by
\begin{equation}
\mathbf{x}^{\prime}\left(  y_{1},y_{2},\cdots,y_{n}\right)  =\int
d\mathbf{x}\Pr\left(  \mathbf{x}|y_{1},y_{2},\cdots,y_{n}\right)
\,\mathbf{x}\label{Eq:ReferenceVector}%
\end{equation}
where $\Pr\left(  \mathbf{x}|y_{1},y_{2},\cdots,y_{n}\right)  $ may be
expressed in terms of $\Pr\left(  \mathbf{x}\right)  $ and $\Pr\left(
y_{1},y_{2},\cdots,y_{n}|\mathbf{x}\right)  $ by using Bayes' theorem in
equation \ref{Eq:Bayes}. The goal now is to minimise the expression for $D$ in
equation \ref{Eq:Objective}\ with respect to the function $\Pr\left(
y_{1},y_{2},\cdots,y_{n}|\mathbf{x}\right)  $. The correct value for
$\mathbf{x}^{\prime}\left(  y_{1},y_{2},\cdots,y_{n}\right)  $ may be
determined by treating it as an unknown parameter that has to be adjusted to
minimise $D$.

$\Pr\left(  y_{1},y_{2},\cdots,y_{n}|\mathbf{x}\right)  $ may be interpreted
as a recognition model which transforms the state of the input layer into (a
probabilistic description of) the state of the output layer, and
$\mathbf{x}^{\prime}\left(  y_{1},y_{2},\cdots,y_{n}\right)  $ may be regarded
as the corresponding generative model that transforms the state of the output
layer into (an approximate reconstruction of) the state of the input layer.%

\sloppypar

There is so much flexibility in the choice of $\Pr\left(  y_{1},y_{2}%
,\cdots,y_{n}|\mathbf{x}\right)  $ (and the corresponding $\mathbf{x}^{\prime
}\left(  y_{1},y_{2},\cdots,y_{n}\right)  $) that even if $D$ is minimised, it
does not necessarily yield an encoded version of the input that is easily
interpretable. One way in which a code can be encouraged to have a simple
interpretation is to force $\mathbf{x}^{\prime}\left(  y_{1},y_{2}%
,\cdots,y_{n}\right)  $ (i.e. the generative model) to be parameterised thus
\cite{Ref:D1D2}
\begin{equation}
\mathbf{x}^{\prime}\left(  y_{1},y_{2},\cdots,y_{n}\right)  =\mathbf{x}%
^{\prime}\left(  y_{1}\right)  +\mathbf{x}^{\prime}\left(  y_{2}\right)
+\cdots+\mathbf{x}^{\prime}\left(  y_{n}\right)
\label{Eq:ReferenceVectorAssumption}%
\end{equation}
which is a (symmetric) superposition of reference vectors $\mathbf{x}^{\prime
}\left(  y\right)  $\ from each neuron $y$ that has been observed to fire. In
this case each neuron has a clearly identifiable contribution to the
reconstruction of the input, which makes it much easier to interpret what each
neuron is doing. \ In this case the $\left\|  \cdots\right\|  ^{2}$ term in
$D$ is symmetric under interchange of the $\left(  y_{1},y_{2},\cdots
,y_{n}\right)  $, so only the symmetric part $S\left[  \Pr\left(  y_{1}%
,y_{2},\cdots,y_{n}|\mathbf{x}\right)  \right]  $ of $\Pr\left(  y_{1}%
,y_{2},\cdots,y_{n}|\mathbf{x}\right)  $ under interchange of the $\left(
y_{1},y_{2},\cdots,y_{n}\right)  $ contributes to $D$, because the symmetric
summation $\sum_{y_{1}=1}^{M}\sum_{y_{2}=1}^{M}\cdots\sum_{y_{n}=1}^{M}\left(
\cdots\right)  $ then removes all non-symmetric contributions.

\bigskip Define the marginal probabilities $\Pr\left(  y_{1}|\mathbf{x}%
\right)  $ and $\Pr\left(  y_{1},y_{2}|\mathbf{x}\right)  $ of the symmetric
part $S\left[  \Pr\left(  y_{1},y_{2},\cdots,y_{n}|\mathbf{x}\right)  \right]
$ of $\Pr\left(  y_{1},y_{2},\cdots,y_{n}|\mathbf{x}\right)  $ under
interchange of the $\left(  y_{1},y_{2},\cdots,y_{n}\right)  $ as
\begin{align}
\Pr\left(  y_{1}|\mathbf{x}\right)   & =\sum_{y_{2},y_{3},y_{4},\cdots
,y_{n}=1}^{M}S\left[  \Pr\left(  y_{1},y_{2},\cdots,y_{n}|\mathbf{x}\right)
\right] \nonumber\\
\Pr\left(  y_{1},y_{2}|\mathbf{x}\right)   & =\sum_{y_{3},y_{4},\cdots
,y_{n}=1}^{M}S\left[  \Pr\left(  y_{1},y_{2},\cdots,y_{n}|\mathbf{x}\right)
\right] \label{Eq:Marginals}%
\end{align}
These marginal probabilities are for the case where $n$ firing events have
potentially been observed, but only the locations of 1 (or 2) firing event(s)
chosen randomly from the total number $n$ have actually been observed, with
the locations of the other $n-1$ (or $n-2$) firing events having been averaged over.

If it is assumed that $\Pr\left(  y_{1}|\mathbf{x}\right)  $ and $\Pr\left(
y_{1},y_{2}|\mathbf{x}\right)  $ are related by
\begin{equation}
\Pr\left(  y_{1},y_{2}|\mathbf{x}\right)  =\Pr\left(  y_{1}|\mathbf{x}\right)
\Pr\left(  y_{2}|\mathbf{x}\right) \label{Eq:PosteriorProbabilityAssumption}%
\end{equation}
then the objective function $D$ has an upper bound $D_{1}+D_{2}$ given by
\cite{Ref:D1D2}
\begin{align}
D  & \leq D_{1}+D_{2}\nonumber\\
D_{1}  & \equiv\frac{2}{n}\int d\mathbf{x}\Pr\left(  \mathbf{x}\right)
\sum_{y=1}^{M}\Pr\left(  y|\mathbf{x}\right)  \left\|  \mathbf{x}%
-\mathbf{x}^{\prime}\left(  y\right)  \right\|  ^{2}\nonumber\\
D_{2}  & \equiv\frac{2\left(  n-1\right)  }{n}\int d\mathbf{x}\Pr\left(
\mathbf{x}\right)  \left\|  \mathbf{x}-\sum_{y=1}^{M}\Pr\left(  y|\mathbf{x}%
\right)  \mathbf{x}^{\prime}\left(  y\right)  \right\|  ^{2}%
\end{align}
Each of the two marginal probabilities in equation \ref{Eq:Marginals}
contributes to a different term in $D_{1}+D_{2}$; $\Pr\left(  y_{1}%
|\mathbf{x}\right)  $ contributes to $D_{1}$, whereas $\Pr\left(  y_{1}%
,y_{2}|\mathbf{x}\right)  $ contributes to $D_{2}$. Informally speaking,
$D_{1}$ measures the information that a single firing event (out of $n$ such
events) contributes to the reconstruction of the input, whereas $D_{2}$
measures the information that pairs of firing events (out of $n$ such events)
contribute to the reconstruction of the input. $D_{1}$ is weighted by a factor
$\frac{1}{n}$ which suppresses the single firing event contribution as
$n\rightarrow\infty$, whereas $D_{2}$ is weighted by a factor $\frac{n-1}{n}$
which suppresses the double firing event contribution as $n\rightarrow1$, as
expected. If only the $D_{1}$ part of the objective function is used (i.e.
$n=1$), then a standard LBG vector quantiser \cite{Ref:LBG} emerges which
approximates the input by a single reference vector $\mathbf{x}^{\prime
}\left(  y\right)  $, whereas if only the $D_{2}$ part of the objective
function is used (i.e. $n\rightarrow\infty$), then the network behaves
essentially as a principal component analyser (PCA) which approximates the
input by a sum of reference vectors $\sum_{y=1}^{M}\Pr\left(  y|\mathbf{x}%
\right)  \mathbf{x}^{\prime}\left(  y\right)  $, where the $\Pr\left(
y|\mathbf{x}\right)  $ are expansion coefficients which sum to unity, and the
$\mathbf{x}^{\prime}\left(  y\right)  $ are basis vectors.

The upper bound $D_{1}+D_{2}$ on $D$ contains LBG encoding and PCA\ encoding
as two limiting cases, and gives a principled way of interpolating between
these extremes. This useful property has been bought at the cost of replacing
$D$ by an upper bound bound $D_{1}+D_{2}$, which will yield only a suboptimal
(from the point of view of $D$) encoder. However, this upper bound can be
expected to be tight in cases where the input manifold can be modelled
accurately using the parameteric form $\mathbf{x}^{\prime}\left(
y_{1}\right)  +\mathbf{x}^{\prime}\left(  y_{2}\right)  +\cdots+\mathbf{x}%
^{\prime}\left(  y_{n}\right)  $. These conditions are well approximated in
images which consist of a discrete number of constituents, each of which may
be represented by an $\mathbf{x}^{\prime}\left(  y\right)  $ for some choice
of $y$. This model fails in situations where two or more constituents are
placed so that they overlap, in which case the image will typically contain
occluded objects, whereas the model assumes that the objects linearly
superpose. Occlusion is not an easy situation to model, so it will be assumed
that the image constituents are sufficiently sparse that they rarely occude
each other.

\section{Stationarity Conditions}

\label{Appendix:Stationarity}The expression for $D_{1}+D_{2}$ (see equation
\ref{Eq:ObjectiveD1D2}) has two types of parameters that need to be optimised:
the reference vectors $\mathbf{x}^{\prime}\left(  y\right)  $ and the
posterior probabilities $\Pr\left(  y|\mathbf{x}\right)  $. In appendix
\ref{Appendix:StationarityX} the stationarity condition for $\mathbf{x}%
^{\prime}\left(  y\right)  $ is derived, and in appendix
\ref{Appendix:StationarityP} the stationarity condition for $\Pr\left(
y|\mathbf{x}\right)  $ is derived, taking into account the constraints
$0\leq\Pr\left(  y|\mathbf{x}\right)  \leq1$ and $\sum_{y=1}^{M}\Pr\left(
y|\mathbf{x}\right)  =1$ which must be satisfied by probabilities.

\subsection{Stationary $\mathbf{x}^{\prime}\left(  y\right)  $}

\label{Appendix:StationarityX}The stationarity condition $\frac{\partial
\left(  D_{1}+D_{2}\right)  }{\partial\mathbf{x}^{\prime}\left(  y\right)
}=0$ for $\mathbf{x}^{\prime}\left(  y\right)  $ was derived in \cite{Ref:WTA}%
. Thus $\frac{\partial\left(  D_{1}+D_{2}\right)  }{\partial\mathbf{x}%
^{\prime}\left(  y\right)  }$ can be written as
\begin{align}
\frac{\partial\left(  D_{1}+D_{2}\right)  }{\partial\mathbf{x}^{\prime}\left(
y\right)  }   =&-\frac{4}{n}\int d\mathbf{x}\Pr\left(  \mathbf{x}\right)
\Pr\left(  y\mathbf{|x}\right) \label{Eq:D1D2Deriv}\\
& \ \times\left(
\begin{array}
[c]{c}%
\mathbf{x-x}^{\prime}\left(  y\right) \\
\\
+\left(  n-1\right)  \sum_{y^{\prime}=1}^{M}\Pr\left(  y^{\prime}%
\mathbf{|x}\right)  \left(  \mathbf{x-x}^{\prime}\left(  y^{\prime}\right)
\right)
\end{array}
\right) \nonumber
\end{align}
and, using Bayes' theorem in the form $\Pr\left(  \mathbf{x|}y\right)
\Pr\left(  y\right)  =\Pr\left(  y|\mathbf{x}\right)  \Pr\left(
\mathbf{x}\right)  $, this yields a matrix equation for the $\mathbf{x}%
^{\prime}\left(  y\right)  $
\begin{equation}
0=\Pr\left(  y\right)  \left(  n\int d\mathbf{x}\Pr\left(  \mathbf{x|}%
y\right)  \,\mathbf{x-}\left(  n-1\right)  \sum_{y^{\prime}=1}^{M}\left(  \int
d\mathbf{x}\Pr\left(  \mathbf{x}|y\right)  \Pr\left(  y^{\prime}%
|\mathbf{x}\right)  \right)  \mathbf{x}^{\prime}\left(  y^{\prime}\right)
-\mathbf{x}^{\prime}\left(  y\right)  \right) \label{Eq:D1D2Stationarity}%
\end{equation}
There are two classes of solution to this stationarity condition,
corresponding to one (or more) of the two factors in equation
\ref{Eq:D1D2Stationarity} being zero.

\begin{enumerate}
\item $\Pr\left(  y\right)  =0$ (the first factor is zero). If the probability
that neuron $y$ fires is zero, then nothing can be deduced about
$\mathbf{x}^{\prime}\left(  y\right)  $, because there is no training data to
explore this neuron's behaviour.

\item $n\int d\mathbf{x}\Pr\left(  \mathbf{x|}y\right)  \,\mathbf{x=}\left(
n-1\right)  \sum_{y^{\prime}=1}^{M}\left(  \int d\mathbf{x}\Pr\left(
\mathbf{x}|y\right)  \Pr\left(  y^{\prime}|\mathbf{x}\right)  \right)
\mathbf{x}^{\prime}\left(  y^{\prime}\right)  +\mathbf{x}^{\prime}\left(
y\right)  $ (the second factor is zero). The solution to this matrix equation
is the required $\mathbf{x}^{\prime}\left(  y\right)  $.
\end{enumerate}

\subsection{Stationary $\Pr\left(  y|\mathbf{x}\right)  $}

\label{Appendix:StationarityP}The stationarity condition $\frac{\delta\left(
D_{1}+D_{2}\right)  }{\delta\log\Pr\left(  y|\mathbf{x}\right)  }$ (with the
normalisation constraint $\sum_{y^{\prime}=1}^{M}\Pr\left(  y^{\prime
}|\mathbf{x}\right)  =1$)\ for $\Pr\left(  y|\mathbf{x}\right)  $
will now be derived.
Thus functionally differentiate $D_{1}+D_{2}$ with
respect to $\log\Pr\left(  y|\mathbf{x}\right)  $, where logarithmic
differentation implicitly imposes the constraint $\Pr\left(  y|\mathbf{x}%
\right)  \geq0$, and use a Lagrange multiplier term $L\equiv\int
d\mathbf{x}^{\prime}\lambda\left(  \mathbf{x}^{\prime}\right)  \sum
_{y^{\prime}=1}^{M}\Pr\left(  y^{\prime}|\mathbf{x}^{\prime}\right)  $ to
impose the normalisation constraint $\sum_{y=1}^{M}\Pr\left(  y|\mathbf{x}%
\right)  =1$ for each $\mathbf{x}$, to obtain
\begin{align}
\frac{\delta\left(  D_{1}+D_{2}-L\right)  }{\delta\log\Pr\left(
y|\mathbf{x}\right)  }   =&\frac{2}{n}\Pr\left(  \mathbf{x}\right)  \Pr\left(
y|\mathbf{x}\right)  \left\|  \mathbf{x}-\mathbf{x}^{\prime}\left(  y\right)
\right\|  ^{2}\nonumber\\
& -\frac{4\left(  n-1\right)  }{n}\Pr\left(  \mathbf{x}\right)  \Pr\left(
y|\mathbf{x}\right) \nonumber\\
& \times\mathbf{x}^{\prime}\left(  y\right)  \cdot\left(  \mathbf{x}%
-\sum_{y=1}^{M}\Pr\left(  y|\mathbf{x}\right)  \,\mathbf{x}^{\prime}\left(
y\right)  \right) \nonumber\\
& -\lambda\left(  \mathbf{x}\right)  \Pr\left(  y|\mathbf{x}\right)
\label{Eq:FunctionalDerivative}%
\end{align}
The stationarity condition implies that $\sum_{y=1}^{M}\Pr\left(
y|\mathbf{x}\right)  \frac{\delta\left(  D_{1}+D_{2}-L\right)  }{\delta
\Pr\left(  y|\mathbf{x}\right)  }=0$, which may be used to determine the
Lagrange multiplier function $\lambda\left(  \mathbf{x}\right)  $. When
$\lambda\left(  \mathbf{x}\right)  $ is substituted back into the stationarity
condition itself, it yields
\begin{align}
0   =&\Pr\left(  \mathbf{x}\right)  \Pr\left(  y|\mathbf{x}\right)
\sum_{y^{\prime}=1}^{M}\left(  \Pr\left(  y^{\prime}|\mathbf{x}\right)
-\delta_{y,y^{\prime}}\right) \nonumber\\
& \times\mathbf{x}^{\prime}\left(  y^{\prime}\right)  \cdot\left(
\frac{\mathbf{x}^{\prime}\left(  y^{\prime}\right)  }{2}-n\mathbf{x+}\left(
n-1\right)  \sum_{y^{\prime\prime}=1}^{M}\Pr\left(  y^{\prime\prime
}|\mathbf{x}\right)  \,\mathbf{x}^{\prime}\left(  y^{\prime\prime}\right)
\right) \label{Eq:StationarityP}%
\end{align}
There are several classes of solution to this stationarity condition,
corresponding to one (or more) of the three factors in equation
\ref{Eq:StationarityP} being zero.

\begin{enumerate}
\item $\Pr\left(  \mathbf{x}\right)  =0$ (the first factor is zero). If the
input PDF is zero at $\mathbf{x}$, then nothing can be deduced about
$\Pr\left(  y|\mathbf{x}\right)  $, because there is no training data to
explore the network's behaviour at this point.

\item $\Pr\left(  y|\mathbf{x}\right)  =0$ (the second factor is zero). This
factor arises from the differentiation with respect to $\log\Pr\left(
y|\mathbf{x}\right)  $, and it ensures that $\Pr\left(  y|\mathbf{x}\right)
<0$ cannot be attained. The singularity in $\log\Pr\left(  y|\mathbf{x}%
\right)  $ when $\Pr\left(  y|\mathbf{x}\right)  =0$ is what causes this
solution to emerge.

\item $\sum_{y^{\prime}=1}^{M}\left(  \Pr\left(  y^{\prime}|\mathbf{x}\right)
-\delta_{y,y^{\prime}}\right)  \mathbf{x}^{\prime}\left(  y^{\prime}\right)
\cdot\left(  \cdots\right)  =0$ (the third factor is zero). The solution to
this equation is a $\Pr\left(  y|\mathbf{x}\right)  $ that has a piecewise
linear dependence on $\mathbf{x}$. This result can be seen to be intuitively
reasonable because $D_{1}+D_{2}$ is of the form $\int d\mathbf{x}\Pr\left(
\mathbf{x}\right)  f\left(  \mathbf{x}\right)  $, where $f\left(
\mathbf{x}\right)  $ is a linear combination of terms of the form
$\mathbf{x}^{i}\Pr\left(  y|\mathbf{x}\right)  ^{j}$ (for $i=0,1,2$ and
$j=0,1,2$), which is a quadratic form in $\mathbf{x}$ (ignoring the
$\mathbf{x}$-dependence of $\Pr\left(  y|\mathbf{x}\right)  $). However, the
terms that appear in this linear combination are such that a $\Pr\left(
y|\mathbf{x}\right)  $ that is a piecewise linear function of $\mathbf{x}$
guarantees that $f\left(  \mathbf{x}\right)  $ is a piecewise linear
combination of terms of the form $\mathbf{x}^{i}$ (for $i=0,1,2$), which is a
quadratic form in $\mathbf{x}$ (the normalisation constraint $\sum_{y=1}%
^{M}\Pr\left(  y|\mathbf{x}\right)  =1$ is used to remove a contribution to
that is potentially quartic in $\mathbf{x}$). Thus a piecewise linear
dependence of $\Pr\left(  y|\mathbf{x}\right)  $ on $\mathbf{x}$ does not lead
to any dependencies on $\mathbf{x}$ that are not already explicitly present in
$D_{1}+D_{2}$. The stationarity condition on $\Pr\left(  y|\mathbf{x}\right)
$ (see equation \ref{Eq:StationarityP})\ then imposes conditions on the
allowed piecewise linearities that $\Pr\left(  y|\mathbf{x}\right)  $ can have.
\end{enumerate}

\section{Simplified Expressions for $D_{1}+D_{2}$}

The expressions for $D_{1}$ and $D_{2}$ (see equation \ref{Eq:ObjectiveD1D2}%
)\ may be simplified in the case of joint encoding and factorial encoding. The
case of joint encoding is derived in appendix \ref{Appendix:Joint}, and the
case of factorial encoding is derived in appendix \ref{Appendix:Factorial}. In
both cases it is assumed that $\mathbf{x}=\left(  \mathbf{x}_{1}%
,\mathbf{x}_{2}\right)  $ and $\Pr\left(  \mathbf{x}_{1},\mathbf{x}%
_{2}\right)  =\Pr\left(  \mathbf{x}_{1}\right)  \Pr\left(  \mathbf{x}%
_{2}\right)  $ where $\Pr\left(  \mathbf{x}_{1}\right)  $ and $\Pr\left(
\mathbf{x}_{2}\right)  $ each define a uniform PDF on the input manifold.

\subsection{Joint Encoding}

\bigskip\label{Appendix:Joint}The expressions for $D_{1}$ and $D_{2}$ may be
simplified in the case of joint encoding, where $\mathbf{x}=\left(
\mathbf{x}_{1},\mathbf{x}_{2}\right)  $, $y=\left(  y_{1},y_{2}\right)  $ for
$1\leq y_{1}\leq\sqrt{M}$ and $1\leq y_{2}\leq\sqrt{M}$. In the following two
derivations of the expressions for $D_{1}$ and $D_{2}$ the steps in the
derivation use exactly the same sequence of manipulations.

The expression for $D_{1}$ is
\begin{align}
D_{1}   =&\frac{2}{n}\int d\mathbf{x}_{1}d\mathbf{x}_{2}\Pr\left(
\mathbf{x}_{1},\mathbf{x}_{2}\right)  \sum_{y_{1}=1}^{\sqrt{M}}\sum_{y_{2}%
=1}^{\sqrt{M}}\Pr\left(  y_{1},y_{2}|\mathbf{x}_{1},\mathbf{x}_{2}\right)
\nonumber\\
& \times\left\|  \left(
\begin{array}
[c]{c}%
\mathbf{x}_{1}\\
\mathbf{x}_{2}%
\end{array}
\right)  -\left(
\begin{array}
[c]{c}%
\mathbf{x}_{1}^{\prime}\left(  y_{1},y_{2}\right) \\
\mathbf{x}_{2}^{\prime}\left(  y_{1},y_{2}\right)
\end{array}
\right)  \right\|  ^{2}%
\end{align}
The assumed properties of $\Pr\left(  \mathbf{x}_{1},\mathbf{x}_{2}\right)  $
imply that $\mathbf{x}_{1}^{\prime}\left(  y_{1},y_{2}\right)  =\mathbf{x}%
_{1}^{\prime}\left(  y_{1}\right)  $ and $\mathbf{x}_{2}^{\prime}\left(
y_{1},y_{2}\right)  =\mathbf{x}_{2}^{\prime}\left(  y_{2}\right)  $, which
gives
\begin{align}
D_{1}   =&\frac{2}{n}\int d\mathbf{x}_{1}d\mathbf{x}_{2}\Pr\left(
\mathbf{x}_{1},\mathbf{x}_{2}\right)  \sum_{y_{1}=1}^{\sqrt{M}}\sum_{y_{2}%
=1}^{\sqrt{M}}\Pr\left(  y_{1},y_{2}|\mathbf{x}_{1},\mathbf{x}_{2}\right)
\nonumber\\
& \times\left(  \left\|  \mathbf{x}_{1}-\mathbf{x}_{1}^{\prime}\left(
y_{1}\right)  \right\|  ^{2}+\left\|  \mathbf{x}_{2}-\mathbf{x}_{2}^{\prime
}\left(  y_{2}\right)  \right\|  ^{2}\right)
\end{align}
Marginalise $\Pr\left(  y_{1},y_{2}|\mathbf{x}_{1},\mathbf{x}_{2}\right)  $
where possible, using that $\sum_{y_{1}=1}^{\sqrt{M}}\Pr\left(  y_{1}%
,y_{2}|\mathbf{x}_{1},\mathbf{x}_{2}\right)  =\Pr\left(  y_{2}|\mathbf{x}%
_{1},\mathbf{x}_{2}\right)  =\Pr\left(  y_{2}|\mathbf{x}_{2}\right)  $ and
$\sum_{y_{2}=1}^{\sqrt{M}}\Pr\left(  y_{1},y_{2}|\mathbf{x}_{1},\mathbf{x}%
_{2}\right)  =\Pr\left(  y_{1}|\mathbf{x}_{1},\mathbf{x}_{2}\right)
=\Pr\left(  y_{1}|\mathbf{x}_{1}\right)  $, to obtain
\begin{equation}
D_{1}=\frac{2}{n}\int d\mathbf{x}_{1}d\mathbf{x}_{2}\Pr\left(  \mathbf{x}%
_{1},\mathbf{x}_{2}\right)  \left(
\begin{array}
[c]{c}%
\sum_{y_{1}=1}^{\sqrt{M}}\Pr\left(  y_{1}|\mathbf{x}_{1}\right)  \left\|
\mathbf{x}_{1}-\mathbf{x}_{1}^{\prime}\left(  y_{1}\right)  \right\|  ^{2}\\
+\sum_{y_{2}=1}^{\sqrt{M}}\Pr\left(  y_{2}|\mathbf{x}_{2}\right)  \left\|
\mathbf{x}_{2}-\mathbf{x}_{2}^{\prime}\left(  y_{2}\right)  \right\|  ^{2}%
\end{array}
\right)
\end{equation}
Marginalise $\Pr\left(  \mathbf{x}_{1},\mathbf{x}_{2}\right)  $ where
possible, using that $\int d\mathbf{x}_{1}\Pr\left(  \mathbf{x}_{1}%
,\mathbf{x}_{2}\right)  =\Pr\left(  \mathbf{x}_{2}\right)  $ and $\int
d\mathbf{x}_{2}\Pr\left(  \mathbf{x}_{1},\mathbf{x}_{2}\right)  =\Pr\left(
\mathbf{x}_{1}\right)  $, to obtain
\begin{align}
D_{1}   =&\frac{2}{n}\int dx_{1}\Pr\left(  \mathbf{x}_{1}\right)  \sum
_{y_{1}=1}^{\sqrt{M}}\Pr\left(  y_{1}|\mathbf{x}_{1}\right)  \left\|
\mathbf{x}_{1}-\mathbf{x}_{1}^{\prime}\left(  y_{1}\right)  \right\|
^{2}\nonumber\\
& +\frac{2}{n}\int d\mathbf{x}_{2}\Pr\left(  \mathbf{x}_{2}\right)
\sum_{y_{2}=1}^{\sqrt{M}}\Pr\left(  y_{2}|\mathbf{x}_{2}\right)  \left\|
\mathbf{x}_{2}-\mathbf{x}_{2}^{\prime}\left(  y_{2}\right)  \right\|  ^{2}%
\end{align}
Because of the assumed symmetry of the solution, these two terms are the same,
which gives
\begin{equation}
D_{1}=\frac{4}{n}\int d\mathbf{x}_{1}\Pr\left(  \mathbf{x}_{1}\right)
\sum_{y_{1}=1}^{\sqrt{M}}\Pr\left(  y_{1}|\mathbf{x}_{1}\right)  \left\|
\mathbf{x}_{1}-\mathbf{x}_{1}^{\prime}\left(  y_{1}\right)  \right\|  ^{2}%
\end{equation}

The expression for $D_{2}$ is
\begin{align}
D_{2}   =&\frac{2\left(  n-1\right)  }{n}\int d\mathbf{x}_{1}d\mathbf{x}%
_{2}\Pr\left(  \mathbf{x}_{1},\mathbf{x}_{2}\right) \nonumber\\
& \times\left\|  \left(
\begin{array}
[c]{c}%
\mathbf{x}_{1}\\
\mathbf{x}_{2}%
\end{array}
\right)  -\sum_{y_{1}=1}^{\sqrt{M}}\sum_{y_{2}=1}^{\sqrt{M}}\Pr\left(
y_{1},y_{2}|\mathbf{x}_{1},\mathbf{x}_{2}\right)  \left(
\begin{array}
[c]{c}%
\mathbf{x}_{1}^{\prime}\left(  y_{1},y_{2}\right) \\
\mathbf{x}_{2}^{\prime}\left(  y_{1},y_{2}\right)
\end{array}
\right)  \right\|  ^{2}%
\end{align}
Use that $\mathbf{x}_{1}^{\prime}\left(  y_{1},y_{2}\right)  =\mathbf{x}%
_{1}^{\prime}\left(  y_{1}\right)  $ and $\mathbf{x}_{2}^{\prime}\left(
y_{1},y_{2}\right)  =\mathbf{x}_{2}^{\prime}\left(  y_{2}\right)  $.
\begin{align}
D_{2}   =&\frac{2\left(  n-1\right)  }{n}\int d\mathbf{x}_{1}d\mathbf{x}%
_{2}\Pr\left(  \mathbf{x}_{1},\mathbf{x}_{2}\right) \nonumber\\
& \times\left(
\begin{array}
[c]{c}%
\left\|  \mathbf{x}_{1}-\sum_{y_{1}=1}^{\sqrt{M}}\sum_{y_{2}=1}^{\sqrt{M}}%
\Pr\left(  y_{1},y_{2}|\mathbf{x}_{1},\mathbf{x}_{2}\right)  \,\mathbf{x}%
_{1}^{\prime}\left(  y_{1}\right)  \right\|  ^{2}\\
+\left\|  \mathbf{x}_{2}-\sum_{y_{1}=1}^{\sqrt{M}}\sum_{y_{2}=1}^{\sqrt{M}}%
\Pr\left(  y_{1},y_{2}|\mathbf{x}_{1},\mathbf{x}_{2}\right)  \,\mathbf{x}%
_{2}^{\prime}\left(  y_{2}\right)  \right\|  ^{2}%
\end{array}
\right)
\end{align}
Use that $\sum_{y_{1}=1}^{\sqrt{M}}\Pr\left(  y_{1},y_{2}|\mathbf{x}%
_{1},\mathbf{x}_{2}\right)  =\Pr\left(  y_{2}|\mathbf{x}_{2}\right)  $ and
$\sum_{y_{2}=1}^{\sqrt{M}}\Pr\left(  y_{1},y_{2}|\mathbf{x}_{1},\mathbf{x}%
_{2}\right)  =\Pr\left(  y_{1}|\mathbf{x}_{1}\right)  $.
\begin{align}
D_{2}   =&\frac{2\left(  n-1\right)  }{n}\int d\mathbf{x}_{1}d\mathbf{x}%
_{2}\Pr\left(  \mathbf{x}_{1},\mathbf{x}_{2}\right) \nonumber\\
& \times\left(
\begin{array}
[c]{c}%
\left\|  \mathbf{x}_{1}-\sum_{y_{1}=1}^{\sqrt{M}}\Pr\left(  y_{1}%
|\mathbf{x}_{1}\right)  \,\mathbf{x}_{1}^{\prime}\left(  y_{1}\right)
\right\|  ^{2}\\
+\left\|  \mathbf{x}_{2}-\sum_{y_{2}=1}^{\sqrt{M}}\Pr\left(  y_{2}%
|\mathbf{x}_{2}\right)  \,\mathbf{x}_{2}^{\prime}\left(  y_{2}\right)
\right\|  ^{2}%
\end{array}
\right)
\end{align}
Marginalise $\Pr\left(  \mathbf{x}_{1},\mathbf{x}_{2}\right)  $.
\begin{align}
D_{2}   =&\frac{2\left(  n-1\right)  }{n}\int d\mathbf{x}_{1}\Pr\left(
\mathbf{x}_{1}\right)  \left\|  \mathbf{x}_{1}-\sum_{y_{1}=1}^{\sqrt{M}}%
\Pr\left(  y_{1}|\mathbf{x}_{1}\right)  \,\mathbf{x}_{1}^{\prime}\left(
y_{1}\right)  \right\|  ^{2}\nonumber\\
& +\frac{2\left(  n-1\right)  }{n}\int d\mathbf{x}_{2}\Pr\left(
\mathbf{x}_{2}\right)  \left\|  \mathbf{x}_{2}-\sum_{y_{2}=1}^{\sqrt{M}}%
\Pr\left(  y_{2}|\mathbf{x}_{2}\right)  \mathbf{\,x}_{2}^{\prime}\left(
y_{2}\right)  \right\|  ^{2}%
\end{align}
Use symmetry$.$%
\begin{equation}
D_{2}=\frac{4\left(  n-1\right)  }{n}\int d\mathbf{x}_{1}\Pr\left(
\mathbf{x}_{1}\right)  \left\|  \mathbf{x}_{1}-\sum_{y_{1}=1}^{\sqrt{M}}%
\Pr\left(  y_{1}|\mathbf{x}_{1}\right)  \,\mathbf{x}_{1}^{\prime}\left(
y_{1}\right)  \right\|  ^{2}%
\end{equation}

These results may be combined to yield finally
\begin{align}
D_{1}+D_{2}   =&\frac{4}{n}\int d\mathbf{x}_{1}\Pr\left(  \mathbf{x}%
_{1}\right)  \sum_{y_{1}=1}^{\sqrt{M}}\Pr\left(  y_{1}|\mathbf{x}_{1}\right)
\left\|  \mathbf{x}_{1}-\mathbf{x}_{1}^{\prime}\left(  y_{1}\right)  \right\|
^{2}\nonumber\\
& +\frac{4\left(  n-1\right)  }{n}\int d\mathbf{x}_{1}\Pr\left(
\mathbf{x}_{1}\right)  \left\|  \mathbf{x}_{1}-\sum_{y_{1}=1}^{\sqrt{M}}%
\Pr\left(  y_{1}|\mathbf{x}_{1}\right)  \,\mathbf{x}_{1}^{\prime}\left(
y_{1}\right)  \right\|  ^{2}%
\end{align}
which has the same form as $D_{1}+D_{2}$ would have had for $\mathbf{x}_{1}%
$-space alone, with the replacement $M\rightarrow\frac{M}{2}$, followed by
multiplication by a factor 2 overall. This implies that the problem of
optimising a joint encoder is trivially related to the problem of optimising
an encoder in the $\mathbf{x}_{1}$-space alone.

\subsection{Factorial Encoding}

\bigskip\label{Appendix:Factorial}The expressions for $D_{1}$ and $D_{2}$ may
be simplified in the case of factorial encoding. In the following two
derivations of the expressions for $D_{1}$ and $D_{2}$, the steps in the
derivation use exactly the same sequence of manipulations, except that $D_{2}
$ has one additional step which separates the contributions inside $\left\|
\cdots\right\|  ^{2}$.

The expression for $D_{1}$ is
\begin{align}
D_{1}   =&\frac{2}{n}\int d\mathbf{x}_{1}d\mathbf{x}_{2}\Pr\left(
\mathbf{x}_{1},\mathbf{x}_{2}\right)  \sum_{y=1}^{M}\Pr\left(  y|\mathbf{x}%
_{1},\mathbf{x}_{2}\right) \nonumber\\
& \times\left\|  \left(
\begin{array}
[c]{c}%
\mathbf{x}_{1}\\
\mathbf{x}_{2}%
\end{array}
\right)  -\left(
\begin{array}
[c]{c}%
\mathbf{x}_{1}^{\prime}\left(  y\right) \\
\mathbf{x}_{2}^{\prime}\left(  y\right)
\end{array}
\right)  \right\|  ^{2}%
\end{align}
Split up $\Pr\left(  y|\mathbf{x}_{1},\mathbf{x}_{2}\right)  $, using that
$\Pr\left(  y|\mathbf{x}_{1},\mathbf{x}_{2}\right)  =\frac{1}{2}\Pr\left(
y|\mathbf{x}_{1}\right)  +\frac{1}{2}\Pr\left(  y|\mathbf{x}_{2}\right)  $,
which gives
\begin{align}
D_{1}   =&\frac{1}{n}\int d\mathbf{x}_{1}d\mathbf{x}_{2}\Pr\left(
\mathbf{x}_{1},\mathbf{x}_{2}\right)  \sum_{y=1}^{M}\left(  \Pr\left(
y|\mathbf{x}_{1}\right)  +\Pr\left(  y|\mathbf{x}_{2}\right)  \right)
\nonumber\\
& \times\left\|  \left(
\begin{array}
[c]{c}%
\mathbf{x}_{1}\\
\mathbf{x}_{2}%
\end{array}
\right)  -\left(
\begin{array}
[c]{c}%
\mathbf{x}_{1}^{\prime}\left(  y\right) \\
\mathbf{x}_{2}^{\prime}\left(  y\right)
\end{array}
\right)  \right\|  ^{2}%
\end{align}
Assume that the input manifold is such that $\mathbf{x}_{1}^{\prime}\left(
y\right)  =\mathbf{0}$ for $\frac{M}{2}+1\leq y\leq M$, and $\mathbf{x}%
_{2}^{\prime}\left(  y\right)  =\mathbf{0}$ for $1\leq y\leq\frac{M}{2}$. Also
use that $\Pr\left(  y|\mathbf{x}_{1}\right)  =0$ for $\frac{M}{2}+1\leq y\leq
M$, and $\Pr\left(  y|\mathbf{x}_{2}\right)  =0$ for $1\leq y\leq\frac{M}{2}$,
to obtain
\begin{align}
D_{1}   =&\frac{1}{n}\int d\mathbf{x}_{1}d\mathbf{x}_{2}\Pr\left(
\mathbf{x}_{1},\mathbf{x}_{2}\right)\sum_{y=1}^{\frac{M}{2}} \Pr\left(  y|\mathbf{x}_{1}\right) \left\|  \left(
\begin{array}
[c]{c}%
\mathbf{x}_{1}\\
\mathbf{x}_{2}%
\end{array}
\right)  -\left(
\begin{array}
[c]{c}%
\mathbf{x}_{1}^{\prime}\left(  y\right) \\
\mathbf{0}%
\end{array}
\right)  \right\|  ^{2}\nonumber\\
& +\frac{1}{n}\int d\mathbf{x}_{1}d\mathbf{x}_{2}\Pr\left(  \mathbf{x}%
_{1},\mathbf{x}_{2}\right)  \sum_{y=\frac{M}{2}+1}^{M}\Pr\left(
y|\mathbf{x}_{2}\right)  \left\|  \left(
\begin{array}
[c]{c}%
\mathbf{x}_{1}\\
\mathbf{x}_{2}%
\end{array}
\right)  -\left(
\begin{array}
[c]{c}%
\mathbf{0}\\
\mathbf{x}_{2}^{\prime}\left(  y\right)
\end{array}
\right)  \right\|  ^{2}%
\end{align}
Because of the assumed symmetry of the solution, these two terms are the same,
which gives
\begin{equation}
D_{1}=\frac{2}{n}\int d\mathbf{x}_{1}d\mathbf{x}_{2}\Pr\left(  \mathbf{x}%
_{1},\mathbf{x}_{2}\right)  \sum_{y=1}^{\frac{M}{2}}\Pr\left(  y|\mathbf{x}%
_{1}\right)  \left(  \left\|  \mathbf{x}_{1}-\mathbf{x}_{1}^{\prime}\left(
y\right)  \right\|  ^{2}+\left\|  \mathbf{x}_{2}\right\|  ^{2}\right)
\end{equation}
Marginalise $\Pr\left(  \mathbf{x}_{1},\mathbf{x}_{2}\right)  $ where
possible, using that $\int d\mathbf{x}_{1}\Pr\left(  \mathbf{x}_{1}%
,\mathbf{x}_{2}\right)  =\Pr\left(  \mathbf{x}_{2}\right)  $ and $\int
d\mathbf{x}_{2}\Pr\left(  \mathbf{x}_{1},\mathbf{x}_{2}\right)  =\Pr\left(
\mathbf{x}_{1}\right)  $, to obtain%

\begin{equation}
D_{1}=\frac{2}{n}\left(
\begin{array}
[c]{c}%
\int d\mathbf{x}_{1}\Pr\left(  \mathbf{x}_{1}\right)  \sum_{y=1}^{\frac{M}{2}%
}\Pr\left(  y|\mathbf{x}_{1}\right)  \left\|  \mathbf{x}_{1}-\mathbf{x}%
_{1}^{\prime}\left(  y\right)  \right\|  ^{2}\\
+\int d\mathbf{x}_{2}\Pr\left(  \mathbf{x}_{2}\right)  \left\|  \mathbf{x}%
_{2}\right\|  ^{2}%
\end{array}
\right)
\end{equation}

The expression for $D_{2}$ is
\begin{equation}
D_{2}=\frac{2\left(  n-1\right)  }{n}\int d\mathbf{x}_{1}d\mathbf{x}_{2}%
\Pr\left(  \mathbf{x}_{1},\mathbf{x}_{2}\right)  \left\|  \left(
\begin{array}
[c]{c}%
\mathbf{x}_{1}\\
\mathbf{x}_{2}%
\end{array}
\right)  -\sum_{y=1}^{M}\Pr\left(  y|\mathbf{x}_{1},\mathbf{x}_{2}\right)
\left(
\begin{array}
[c]{c}%
\mathbf{x}_{1}^{\prime}\left(  y\right) \\
\mathbf{x}_{2}^{\prime}\left(  y\right)
\end{array}
\right)  \right\|  ^{2}%
\end{equation}
Use that $\Pr\left(  y|\mathbf{x}_{1},\mathbf{x}_{2}\right)  =\frac{1}{2}%
\Pr\left(  y|\mathbf{x}_{1}\right)  +\frac{1}{2}\Pr\left(  y|\mathbf{x}%
_{2}\right)  $.
\begin{align}
D_{2}   =&\frac{2\left(  n-1\right)  }{n}\int d\mathbf{x}_{1}d\mathbf{x}%
_{2}\Pr\left(  \mathbf{x}_{1},\mathbf{x}_{2}\right) \nonumber\\
& \times\left\|  \left(
\begin{array}
[c]{c}%
\mathbf{x}_{1}\\
\mathbf{x}_{2}%
\end{array}
\right)  -\frac{1}{2}\sum_{y=1}^{M}\left(  \Pr\left(  y|\mathbf{x}_{1}\right)
+\Pr\left(  y|\mathbf{x}_{2}\right)  \right)  \left(
\begin{array}
[c]{c}%
\mathbf{x}_{1}^{\prime}\left(  y\right) \\
\mathbf{x}_{2}^{\prime}\left(  y\right)
\end{array}
\right)  \right\|  ^{2}%
\end{align}
Separate the contributions from the upper and lower components inside
$\left\|  \cdots\right\|  ^{2}$, to obtain
\begin{align}
D_{2}   =&\frac{2\left(  n-1\right)  }{n}\int d\mathbf{x}_{1}d\mathbf{x}%
_{2}\Pr\left(  \mathbf{x}_{1},\mathbf{x}_{2}\right) \nonumber\\
& \times\left\|
\begin{array}
[c]{c}%
\left(
\begin{array}
[c]{c}%
\mathbf{x}_{1}\\
\mathbf{x}_{2}%
\end{array}
\right)  -\frac{1}{2}\sum_{y=1}^{\frac{M}{2}}\Pr\left(  y|\mathbf{x}%
_{1}\right)  \left(
\begin{array}
[c]{c}%
\mathbf{x}_{1}^{\prime}\left(  y\right) \\
\mathbf{0}%
\end{array}
\right) \\
-\frac{1}{2}\sum_{y=\frac{M}{2}}^{M}\Pr\left(  y|\mathbf{x}_{2}\right)
\left(
\begin{array}
[c]{c}%
\mathbf{0}\\
\mathbf{x}_{2}^{\prime}\left(  y\right)
\end{array}
\right)
\end{array}
\right\|  ^{2}%
\end{align}
Use that $\mathbf{x}_{1}^{\prime}\left(  y\right)  =\mathbf{0}$ for $\frac
{M}{2}+1\leq y\leq M$, and $\mathbf{x}_{2}^{\prime}\left(  y\right)
=\mathbf{0}$ for $1\leq y\leq\frac{M}{2}$. Also use that $\Pr\left(
y|\mathbf{x}_{1}\right)  =0$ for $\frac{M}{2}+1\leq y\leq M$, and $\Pr\left(
y|\mathbf{x}_{2}\right)  =0$.
\begin{align}
D_{2}   =&\frac{2\left(  n-1\right)  }{n}\int d\mathbf{x}_{1}d\mathbf{x}%
_{2}\Pr\left(  \mathbf{x}_{1},\mathbf{x}_{2}\right)  \left\|  \mathbf{x}%
_{1}-\frac{1}{2}\sum_{y=1}^{\frac{M}{2}}\Pr\left(  y|\mathbf{x}_{1}\right)
\mathbf{x}_{1}^{\prime}\left(  y\right)  \right\|  ^{2}\nonumber\\
& +\frac{2\left(  n-1\right)  }{n}\int d\mathbf{x}_{1}d\mathbf{x}_{2}%
\Pr\left(  \mathbf{x}_{1},\mathbf{x}_{2}\right)  \left\|  \mathbf{x}_{2}%
-\frac{1}{2}\sum_{y=\frac{M}{2}}^{M}\Pr\left(  y|\mathbf{x}_{2}\right)
\mathbf{x}_{2}^{\prime}\left(  y\right)  \right\|  ^{2}%
\end{align}
Use symmetry.
\begin{equation}
D_{2}=\frac{4\left(  n-1\right)  }{n}\int d\mathbf{x}_{1}d\mathbf{x}_{2}%
\Pr\left(  \mathbf{x}_{1},\mathbf{x}_{2}\right)  \left\|  \mathbf{x}_{1}%
-\frac{1}{2}\sum_{y=1}^{\frac{M}{2}}\Pr\left(  y|\mathbf{x}_{1}\right)
\mathbf{x}_{1}^{\prime}\left(  y\right)  \right\|  ^{2}%
\end{equation}
Marginalise $\Pr\left(  \mathbf{x}_{1},\mathbf{x}_{2}\right)  $.
\begin{equation}
D_{2}=\frac{4\left(  n-1\right)  }{n}\int d\mathbf{x}_{1}\Pr\left(
\mathbf{x}_{1}\right)  \left\|  \mathbf{x}_{1}-\frac{1}{2}\sum_{y=1}^{\frac
{M}{2}}\Pr\left(  y|\mathbf{x}_{1}\right)  \mathbf{x}_{1}^{\prime}\left(
y\right)  \right\|  ^{2}%
\end{equation}

These results may be combined to yield finally
\begin{align}
D_{1}+D_{2}   =&\frac{2}{n}\int d\mathbf{x}_{2}\Pr\left(  \mathbf{x}%
_{2}\right)  \left\|  \mathbf{x}_{2}\right\|  ^{2}\nonumber\\
& +\frac{2}{n}\int d\mathbf{x}_{1}\Pr\left(  \mathbf{x}_{1}\right)  \sum
_{y=1}^{\frac{M}{2}}\Pr\left(  y|\mathbf{x}_{1}\right)  \left\|
\mathbf{x}_{1}-\mathbf{x}_{1}^{\prime}\left(  y\right)  \right\|
^{2}\nonumber\\
& +\frac{4\left(  n-1\right)  }{n}\int d\mathbf{x}_{1}\Pr\left(
\mathbf{x}_{1}\right)  \left\|  \mathbf{x}_{1}-\frac{1}{2}\sum_{y=1}^{\frac
{M}{2}}\Pr\left(  y|\mathbf{x}_{1}\right)  \mathbf{x}_{1}^{\prime}\left(
y\right)  \right\|  ^{2}%
\end{align}

The stationarity conditions may be derived from this expression for the
factorial encoding version of $D_{1}+D_{2}$. The stationarity condition w.r.t.
$\Pr\left(  y|\mathbf{x}_{1}\right)  $ is
\begin{equation}
\sum_{y^{\prime}=1}^{\frac{M}{2}}\left(  \Pr\left(  y^{\prime}|\mathbf{x}%
_{1}\right)  -\delta_{y,y^{\prime}}\right)  \mathbf{x}_{1}^{\prime}\left(
y^{\prime}\right)  \cdot\left(  \frac{1}{2}\mathbf{x}_{1}^{\prime}\left(
y^{\prime}\right)  -n\,\mathbf{x}_{1}+\frac{n-1}{2}\sum_{y^{\prime\prime}%
=1}^{\frac{M}{2}}\Pr\left(  y^{\prime\prime}|\mathbf{x}_{1}\right)
\mathbf{x}_{1}^{\prime}\left(  y^{\prime\prime}\right)  \right)  =0
\end{equation}
and the stationarity condition w.r.t. $\mathbf{x}_{1}^{\prime}\left(
y\right)  $ is
\begin{equation}
n\int d\mathbf{x}_{1}\Pr\left(  \mathbf{x}_{1}|y\right)  \mathbf{x}%
_{1}=\mathbf{x}_{1}^{\prime}\left(  y\right)  +\frac{n-1}{2}\int
d\mathbf{x}_{1}\Pr\left(  \mathbf{x}_{1}|y\right)  \sum_{y^{\prime}=1}%
^{\frac{M}{2}}\Pr\left(  y^{\prime}|\mathbf{x}_{1}\right)  \mathbf{x}%
_{1}^{\prime}\left(  y^{\prime}\right)
\end{equation}
Both of these stationarity conditions can be obtained from the standard ones
by making the replacements $\left(  n-1\right)  \sum_{y^{\prime}=1}^{M}%
\Pr\left(  y^{\prime}|\mathbf{x}_{1}\right)  \mathbf{x}_{1}^{\prime}\left(
y^{\prime}\right)  \rightarrow\frac{n-1}{2}\sum_{y^{\prime}=1}^{M}\Pr\left(
y^{\prime}|\mathbf{x}_{1}\right)  \mathbf{x}_{1}^{\prime}\left(  y^{\prime
}\right)  $ and $M\rightarrow\frac{M}{2}$.

\section{Minimise $D_{1}+D_{2}$}

The expression for $D_{1}+D_{2}$ needs to be minimised with respect to the
reference vectors $\mathbf{x}^{\prime}\left(  y\right)  $ and the posterior
probabilities $\Pr\left(  y|\mathbf{x}\right)  $. There are four cases to
consider, which are various combinations of circular/toroidal input manifold
(appendices \ref{Appendix:CircularManifold2} and
\ref{Appendix:CircularManifold3}/appendices \ref{Appendix:ToroidalManifold2}
and \ref{Appendix:ToroidalManifold3}) and two/three overlapping posterior
probabilities (appendices \ref{Appendix:CircularManifold2}
and\ \ref{Appendix:ToroidalManifold2}/appendices
\ref{Appendix:CircularManifold3} and \ref{Appendix:ToroidalManifold3}). For a
toroidal manifold it is not necessary to consider the case of joint encoding,
because it is directly related to encoding a circular manifold, which is dealt
with in appendices \ref{Appendix:CircularManifold2} and
\ref{Appendix:CircularManifold3}.

\subsection{Circular Manifold: 2 Overlapping Posterior Probabilities}

\label{Appendix:CircularManifold2}For $0\leq s\leq\frac{\pi}{M}$ the
functional form of $p\left(  \theta\right)  $ that ensures a piecewise linear
$\Pr\left(  y|\mathbf{x}\right)  $ is
\begin{equation}
p\left(  \theta\right)  =\left\{
\begin{array}
[c]{ll}%
1 & 0\leq\left|  \theta\right|  \leq\frac{\pi}{M}-s\\
f\left(  \theta\right)  & \frac{\pi}{M}-s\leq\left|  \theta\right|  \leq
\frac{\pi}{M}+s\\
0 & \left|  \theta\right|  \geq\frac{\pi}{M}+s
\end{array}
\right.
\end{equation}
where $f\left(  \theta\right)  =a+b\cos\theta+c\sin\left|  \theta\right|  $.
Continuity of $p\left(  \theta\right)  $ gives $f\left(  \frac{\pi}%
{M}-s\right)  =1$ and $f\left(  \frac{\pi}{M}+s\right)  =0$. Normalisation of
$p\left(  \theta\right)  $ in the interval $\frac{\pi}{M}-s\leq\theta\leq
\frac{\pi}{M}+s$ requires that $f\left(  \theta\right)  +f\left(  \frac{2\pi
}{M}-\theta\right)  =1$ . These yield $f\left(  \theta\right)  $ in the form
\begin{equation}
f\left(  \theta\right)  =\frac{1}{2}+\frac{1}{2}\frac{\sin\left(  \frac{\pi
}{M}-\theta\right)  }{\sin s}%
\end{equation}
$D_{1}+D_{2}$ must be stationary w.r.t. variation of $p\left(  \theta\right)
$ in the interval $\frac{\pi}{M}-s\leq\theta\leq\frac{\pi}{M}+s$, which yields
the condition
\begin{align}
0   =&r\csc^{2}s\,\,\sin\left(  \frac{\pi}{M}\right)  \sin\left(  \frac{\pi
}{M}-\theta\right)  \left(  \sin s-\sin\left(  \frac{\pi}{M}-\theta\right)
\right) \nonumber\\
& \times\left(  n\sin s-\left(  n-1\right)  \,r\sin\left(  \frac{\pi}%
{M}\right)  \right)  \,\,
\end{align}
which gives the optimum solution for $r$ as
\begin{equation}
r=\frac{n}{n-1}\frac{\sin s}{\sin\left(  \frac{\pi}{M}\right)  }%
\end{equation}
$D_{1}+D_{2}$ must be stationary w.r.t. variation of $r$. This yields a
transcendental equation that must be satisfied by the optimum solution for $s
$ as
\begin{equation}
\frac{\sin s}{\sin\left(  \frac{\pi}{M}\right)  }-\frac{n-1}{n}\frac{M}{\pi
}\sin\left(  \frac{\pi}{M}\right)  \left(  \cos s+s\sin s\right)  =0
\end{equation}
$D_{1}$ and $D_{2}$ may be written out in full as (using $\mathbf{n}\left(
\theta\right)  \equiv\left(  \cos\theta,\sin\theta\right)  $)
\begin{equation}
D_{1}=\frac{2M}{n\,\pi}\left(
\begin{array}
[c]{c}%
\int_{0}^{\frac{\pi}{M}-s}d\theta\left\|  \mathbf{n}\left(  \theta\right)
-r\,\mathbf{n}\left(  0\right)  \right\|  ^{2}\\
+\int_{\frac{\pi}{M}-s}^{\frac{\pi}{M}}d\theta\,f\left(  \theta\right)
\left\|  \mathbf{n}\left(  \theta\right)  -r\,\mathbf{n}\left(  0\right)
\right\|  ^{2}\\
+\int_{\frac{\pi}{M}-s}^{\frac{\pi}{M}}d\theta\,f\left(  \frac{2\pi}{M}%
-\theta\right)  \left\|  \mathbf{n}\left(  \theta\right)  -r\,\mathbf{n}%
\left(  \frac{2\pi}{M}\right)  \right\|  ^{2}%
\end{array}
\right)
\end{equation}%
\begin{equation}
D_{2}=\frac{2\left(  n-1\right)  M}{n\,\pi}\left(
\begin{array}
[c]{c}%
\int_{0}^{\frac{\pi}{M}-s}d\theta\left\|  \mathbf{n}\left(  \theta\right)
-r\,\mathbf{n}\left(  0\right)  \right\|  ^{2}\\
+\int_{\frac{\pi}{M}-s}^{\frac{\pi}{M}}d\theta\left\|
\begin{array}
[c]{c}%
\mathbf{n}\left(  \theta\right)  -r\,f\left(  \theta\right)  \,\mathbf{n}%
\left(  0\right) \\
-r\,f\left(  \frac{2\pi}{M}-\theta\right)  \mathbf{n}\left(  \frac{2\pi}%
{M}\right)
\end{array}
\right\|  ^{2}%
\end{array}
\right)
\end{equation}
The optimum $f\left(  \theta\right)  $ and $r$ may be substituted into
$D_{1}+D_{2}$, the integrations evaluated, and then the condition that the
optimum $s$ must satisfy may be used to simplify the result, to yield the
minimum $D_{1}+D_{2}$ as
\begin{equation}
D_{1}+D_{2}=2-\frac{n}{n-1}\frac{M}{2\pi}\left(  2s+\sin\left(  2s\right)
\right)
\end{equation}

\subsection{Circular Manifold: 3 Overlapping Posterior Probabilities}

\label{Appendix:CircularManifold3}For $\frac{\pi}{M}\leq s\leq\frac{2\pi}{M}$
the functional form of $p\left(  \theta\right)  $ that ensures a piecewise
linear $\Pr\left(  y|\mathbf{x}\right)  $ is
\begin{equation}
p\left(  \theta\right)  =\left\{
\begin{array}
[c]{ll}%
f_{1}\left(  \theta\right)  & 0\leq\left|  \theta\right|  \leq-\frac{\pi}%
{M}+s\\
f_{2}\left(  \theta\right)  & -\frac{\pi}{M}+s\leq\left|  \theta\right|
\leq\frac{3\pi}{M}-s\\
f_{3}\left(  \theta\right)  & \frac{3\pi}{M}-s\leq\left|  \theta\right|
\leq\frac{\pi}{M}+s\\
0 & \left|  \theta\right|  \geq\frac{\pi}{M}+s
\end{array}
\right.
\end{equation}
where $f_{i}\left(  \theta\right)  =a_{i}+b_{i}\cos\theta+c_{i}\sin\left|
\theta\right|  $ for $i=1,2,3$. Continuity of $p\left(  \theta\right)  $ gives
$f_{1}\left(  -\frac{\pi}{M}+s\right)  =f_{2}\left(  -\frac{\pi}{M}+s\right)
$, $f_{2}\left(  \frac{3\pi}{M}-s\right)  =f_{3}\left(  \frac{3\pi}%
{M}-s\right)  $ and $f_{3}\left(  \frac{2\pi}{M}+s\right)  =0$. Normalisation
of $p\left(  \theta\right)  $ in the interval $0\leq\theta\leq-\frac{\pi}%
{M}+s$ requires that $f_{1}\left(  \theta\right)  +f_{3}\left(  \frac{2\pi}%
{M}+\theta\right)  +f_{3}\left(  \frac{2\pi}{M}-\theta\right)  =1$, and
normalisation of $p\left(  \theta\right)  $ in the interval $-\frac{\pi}%
{M}+s\leq\theta\leq\frac{3\pi}{M}-s$ requires that $f_{2}\left(
\theta\right)  +f_{2}\left(  \frac{2\pi}{M}-\theta\right)  =1$. These
conditions may be used to eliminate all but a pair of parameters in the
$f_{i}\left(  \theta\right)  $, which may thus be written in the form
\begin{align}
f_{1}\left(  \theta\right)    =&\frac{1}{2}\cos\left(  \theta\right)
\sec\left(  \frac{\pi}{M}-s\right) \nonumber\\
& +a_{1}\left(  1-\cos\left(  \theta\right)  \ \sec\left(  \frac{\pi}%
{M}-s\right)  \right) \nonumber\\
& +b_{2}\cos\left(  \theta\right)  \csc\left(  \frac{\pi}{M}\right)
\sin\left(  \frac{2\pi}{M}-s\right)  \sec\left(  \frac{\pi}{M}-s\right)
\nonumber\\
f_{2}\left(  \theta\right)    =&\frac{1}{2}+b_{2}\ \left(  \cos\left(
\theta\right)  -\cot\left(  \frac{\pi}{M}\right)  \ \sin\left(  \theta\right)
\right) \nonumber\\
f_{3}\left(  \theta\right)    =&\frac{1}{2}\left(  1-\csc\left(  \frac{2\pi
}{M}-2s\right)  \sin\left(  \frac{3\pi}{M}-s-\theta\right)  \right)
\nonumber\\
& +\frac{1}{2}a_{1}\left(  \cos\left(  \frac{2\pi}{M}-\theta\right)
\sec\left(  \frac{\pi}{M}-s\right)  -1\right) \nonumber\\
& +b_{2}\csc\left(  \frac{\pi}{M}\right)  \csc\left(  \frac{2\pi}%
{M}-2s\right)  \sin\left(  \frac{2\pi}{M}-s\right)  \sin\left(  \frac{\pi}%
{M}+s-\theta\right)
\end{align}
$D_{1}+D_{2}$ must be stationary w.r.t. variation of $p\left(  \theta\right)
$ in each of the 3 intervals $0\leq\theta\leq-\frac{\pi}{M}+s$ (interval 1),
$-\frac{\pi}{M}+s\leq\theta\leq\frac{3\pi}{M}-s$ (interval 2), and $\frac
{3\pi}{M}-s\leq\theta\leq\frac{\pi}{M}+s$ (interval 3). The Fourier transform
w.r.t. $\theta$ of each of these 3 stationarity conditions has 5 terms with
basis functions $\left(  1,\cos\theta,\sin\theta,\cos2\theta,\sin
2\theta\right)  $, and each of the total of 15 Fourier coefficients must be
zero. There are only 3 free parameters $a_{1}$, $b_{2}$ and $r$, so only 3 of
the 15 are actually independent; the particular 3 that are used are selected
on the basis of ease of solution for the free parameters $a_{1}$, $b_{2}$ and
$r$. The coefficient of the $\cos2\theta$ term in interval 2 yields
\begin{equation}
\ b_{2}\,r\,\left(  n+2\,b_{2}\,r-2\,b_{2}r\,n\right)  \cos\left(  \frac{2\pi
}{M}\right)  =0
\end{equation}
which has the solution
\begin{equation}
b_{2}=\frac{n}{2\,(n-1)\,r}%
\end{equation}
which may be substituted back into the coefficient of the $\cos\theta$ term in
interval 1 to yield
\begin{align}
0   =&r\sec\left(  \frac{\pi}{M}-s\right)  \sin\left(  \frac{\pi}{M}\right)
\nonumber\\
&\times \left(
\begin{array}
[c]{c}%
\left(  n-1\right)  \,\left(  -6\,a_{1}^{2}+7\,a_{1}-2\right)  \,r\sin\left(
\frac{\pi}{M}\right) \\
+\left(  n-1\right)  \,\left(  2\,a_{1}^{2}-3\,a_{1}+1\right)  \,r\sin\left(
\frac{3\pi}{M}\right) \\
+n\,\left(  a_{1}\sin\left(  \frac{2\pi}{M}-s\right)  +\left(  1-a_{1}\right)
\sin\left(  \frac{4\pi}{M}-s\right)  \right)
\end{array}
\right)
\end{align}
and also substituted back into the coefficient of the $\sin\theta$ term in
interval 3 to yield
\begin{align}
0   =&r\,\cos\left(  \frac{\pi}{M}\right)  \,\csc\left(  \frac{\pi}%
{M}-s\right)  \,\sec\left(  \frac{\pi}{M}-s\right)  \,\sin^{2}\left(
\frac{\pi}{M}\right)  \nonumber\\
& \times\left(
\begin{array}
[c]{c}%
-\left(  n-1\right)  \,r\,\left(
\begin{array}
[c]{c}%
-2\,a_{1}\left(  3\,a_{1}-2\right)  \cos\left(  \frac{2\pi}{M}-s\right) \\
-2\,\left(  a_{1}-1\right)  \,\,a_{1}\cos\left(  \frac{2\pi}{M}+s\right) \\
+\left(  1-2a_{1}+2\,a_{1}^{2}\right)  \,\cos\left(  \frac{4\pi}{M}-s\right)
\\
+\left(  1-4\,a_{1}+6\,a_{1}^{2}\right)  \,\cos\left(  s\right)
\end{array}
\right) \\
+n\,\left(
\begin{array}
[c]{c}%
\left(  a_{1}+1\right)  \cos\left(  \frac{\pi}{M}\right)  -\left(
a_{1}-1\right)  \cos\left(  \frac{3\pi}{M}\right) \\
-2\,a_{1}\sin\left(  \frac{\pi}{M}\right)  \sin\left(  \frac{4\pi}%
{M}-2\,s\right)
\end{array}
\right)
\end{array}
\right)
\end{align}
These two conditions may be solved for $a_{1}$ and $r$ to yield
\begin{equation}
a_{1}=\frac{\cos\left(  \frac{2\pi}{M}\right)  }{\cos\left(  \frac{2\pi}%
{M}\right)  -1}%
\end{equation}
and
\begin{equation}
r=\frac{n}{n-1}\cos\left(  \frac{2\pi}{M}-s\right)  \,\sec\left(  \frac{\pi
}{M}\right)
\end{equation}
The solutions for $a_{1}$ and $b_{2}$ may be substituted back into the
expressions for the $f_{i}\left(  \theta\right)  $ to reduce them to the form
\begin{align}
f_{1}\left(  \theta\right)   & =-\frac{1}{4}\left(  \cos\left(  \frac{4\pi}%
{M}-s\right)  +\cos s-2\cos\left(  \frac{\pi}{M}\right)  \cos\theta\right)
\csc^{2}\left(  \frac{\pi}{M}\right)  \sec\left(  \frac{2\pi}{M}-s\right)
\nonumber\\
f_{2}\left(  \theta\right)   & =\frac{1}{2}\left(  \cot\left(  \frac{\pi}%
{M}\right)  \sec\left(  \frac{2\pi}{M}-s\right)  \sin\left(  \frac{\pi}%
{M}-\theta\right)  +1\right) \nonumber\\
f_{3}\left(  \theta\right)   & =-\frac{1}{4}\csc^{2}\left(  \frac{\pi}%
{M}\right)  \left(  \cos\left(  \frac{3\pi}{M}-\theta\right)  \sec\left(
\frac{2\pi}{M}-s\right)  -1\right)
\end{align}
$D_{1}+D_{2}$ must be stationary w.r.t. variation of $r$. This yields a
transcendental equation that must be satisfied by the optimum solution for $s
$ as
\begin{equation}
\frac{1}{n}\frac{\cos\left(  \frac{2\pi}{M}-s\right)  }{\cos\left(  \frac{\pi
}{M}\right)  }-\frac{n-1}{n}\frac{M}{\pi}\cos\left(  \frac{\pi}{M}\right)
\left(  \sin\left(  \frac{2\pi}{M}-s\right)  -\left(  \frac{2\pi}{M}-s\right)
\cos\left(  \frac{2\pi}{M}-s\right)  \right)  =0
\end{equation}
$D_{1}$ and $D_{2}$ may be written out in full as
\begin{equation}
D_{1}=\frac{M}{n\,\pi}\left(
\begin{array}
[c]{c}%
\int_{0}^{-\frac{\pi}{M}+s}d\theta\,\left(
\begin{array}
[c]{c}%
f_{1}\left(  \theta\right)  \left\|  \mathbf{n}\left(  \theta\right)
-r\,\mathbf{n}\left(  0\right)  \right\|  ^{2}\\
+f_{3}\left(  \frac{2\pi}{M}-\theta\right)  \left\|  \mathbf{n}\left(
\theta\right)  -r\,\mathbf{n}\left(  \frac{2\pi}{M}\right)  \right\|  ^{2}\\
+f_{3}\left(  \frac{2\pi}{M}+\theta\right)  \left\|  \mathbf{n}\left(
\theta\right)  -r\,\mathbf{n}\left(  -\frac{2\pi}{M}\right)  \right\|  ^{2}%
\end{array}
\right) \\
+\int_{-\frac{\pi}{M}+s}^{\frac{3\pi}{M}-s}d\theta\,\left(
\begin{array}
[c]{c}%
f_{2}\left(  \theta\right)  \left\|  \mathbf{n}\left(  \theta\right)
-r\,\mathbf{n}\left(  0\right)  \right\|  ^{2}\\
+f_{2}\left(  \frac{2\pi}{M}-\theta\right)  \left\|  \mathbf{n}\left(
\theta\right)  -r\,\mathbf{n}\left(  \frac{2\pi}{M}\right)  \right\|  ^{2}%
\end{array}
\right) \\
+\int_{\frac{3\pi}{M}-s}^{\frac{2\pi}{M}}d\theta\,\left(
\begin{array}
[c]{c}%
f_{3}\left(  \theta\right)  \left\|  \mathbf{n}\left(  \theta\right)
-r\,\mathbf{n}\left(  0\right)  \right\|  ^{2}\\
+f_{1}\left(  \frac{2\pi}{M}-\theta\right)  \left\|  \mathbf{n}\left(
\theta\right)  -r\,\mathbf{n}\left(  \frac{2\pi}{M}\right)  \right\|  ^{2}\\
+f_{3}\left(  \frac{4\pi}{M}-\theta\right)  \left\|  \mathbf{n}\left(
\theta\right)  -r\,\mathbf{n}\left(  \frac{4\pi}{M}\right)  \right\|  ^{2}%
\end{array}
\right)
\end{array}
\right)
\end{equation}%
\begin{equation}
D_{2}=\frac{\left(  n-1\right)  M}{n\,\pi}\left(
\begin{array}
[c]{c}%
\int_{0}^{-\frac{\pi}{M}+s}d\theta\left\|
\begin{array}
[c]{c}%
\mathbf{n}\left(  \theta\right)  -f_{1}\left(  \theta\right)  \,r\,\mathbf{n}%
\left(  0\right) \\
-f_{3}\left(  \frac{2\pi}{M}-\theta\right)  \,r\,\mathbf{n}\left(  \frac{2\pi
}{M}\right) \\
-f_{3}\left(  \frac{2\pi}{M}+\theta\right)  \,r\,\mathbf{n}\left(  -\frac
{2\pi}{M}\right)
\end{array}
\right\|  ^{2}\\
+\int_{-\frac{\pi}{M}+s}^{\frac{3\pi}{M}-s}d\theta\,\left\|
\begin{array}
[c]{c}%
\mathbf{n}\left(  \theta\right)  -f_{2}\left(  \theta\right)  \,r\,\mathbf{n}%
\left(  0\right) \\
-f_{2}\left(  \frac{2\pi}{M}-\theta\right)  \,r\,\mathbf{n}\left(  \frac{2\pi
}{M}\right)
\end{array}
\right\|  ^{2}\\
+\int_{\frac{3\pi}{M}-s}^{\frac{2\pi}{M}}d\theta\,\left\|
\begin{array}
[c]{c}%
\mathbf{n}\left(  \theta\right)  -f_{3}\left(  \theta\right)  \,r\,\mathbf{n}%
\left(  0\right) \\
-f_{1}\left(  \frac{2\pi}{M}-\theta\right)  \,r\,\mathbf{n}\left(  \frac{2\pi
}{M}\right) \\
-f_{3}\left(  \frac{4\pi}{M}-\theta\right)  \,r\,\mathbf{n}\left(  \frac{4\pi
}{M}\right)
\end{array}
\right\|  ^{2}%
\end{array}
\right)
\end{equation}
The optimum $f_{i}\left(  \theta\right)  $ and $r$ may be substituted into
$D_{1}+D_{2}$, the integrations evaluated, and then the condition that the
optimum $s$ must satisfy may be used to simplify the result, to yield the
minimum $D_{1}+D_{2}$ as
\begin{align}
D_{1}+D_{2}   =&\frac{n\,\left(  \left(  n-1\right)  \left(  2\frac{n-2}%
{n}-\frac{M}{\pi}\,s\right)  -\sec^{2}\left(  \frac{\pi}{M}\right)  \right)
}{2\left(  n-1\right)  ^{2}}\nonumber\\
& -\frac{n\,\left(  \left(  n-1\right)  \left(  2-\frac{M}{\pi}\,s\right)
+\sec^{2}\left(  \frac{\pi}{M}\right)  \right)  }{2\left(  n-1\right)  ^{2}%
}\cos\left(  \frac{4\pi}{M}-2s\right)
\end{align}

\subsection{Toroidal Manifold: 2 Overlapping Posterior Probabilities}

\label{Appendix:ToroidalManifold2}For $0\leq s\leq\frac{2\pi}{M}$ the
functional form of $p\left(  \theta\right)  $ may be obtained directly from
the circular case with the replacement $M\rightarrow\frac{M}{2}$, so that
\begin{align}
p\left(  \theta\right)   & =\left\{
\begin{array}
[c]{ll}%
1 & 0\leq\left|  \theta\right|  \leq\frac{2\pi}{M}-s\\
f\left(  \theta\right)  & \frac{2\pi}{M}-s\leq\left|  \theta\right|  \leq
\frac{2\pi}{M}+s\\
0 & \left|  \theta\right|  \geq\frac{2\pi}{M}+s
\end{array}
\right. \\
f\left(  \theta\right)   & =\frac{1}{2}+\frac{1}{2}\frac{\sin\left(
\frac{2\pi}{M}-\theta\right)  }{\sin s}%
\end{align}
$D_{1}+D_{2}$ must be stationary w.r.t. variation of $p\left(  \theta\right)
$ in the interval $\frac{2\pi}{M}-s\leq\theta\leq\frac{2\pi}{M}+s$, which
yields the condition
\begin{align}
0   =&r\csc^{2}s\,\sin\left(  \frac{2\pi}{M}\right)  \,\sin\left(  \frac{2\pi
}{M}-\theta\right)  \left(  \sin s-\sin\left(  \frac{2\pi}{M}-\theta\right)
\right) \nonumber\\
& \times\left(  2\,n\sin s-\left(  n-1\right)  \,r\sin\left(  \frac{2\pi}%
{M}\right)  \right)
\end{align}
which has the same form as the circular case with the replacements
$M\rightarrow\frac{M}{2}$ and $n\rightarrow\frac{2n}{n+1}$, which gives the
optimum solution for $r$ as
\begin{equation}
r=\frac{2n}{n-1}\frac{\sin s}{\sin\left(  \frac{2\pi}{M}\right)  }%
\end{equation}
$D_{1}+D_{2}$ must be stationary w.r.t. variation of $r$. This yields a
transcendental equation that must be satisfied by the optimum solution for $s
$ as
\begin{equation}
\frac{\sin s}{\sin\left(  \frac{2\pi}{M}\right)  }-\frac{n-1}{n+1}\frac
{M}{2\pi}\sin\left(  \frac{2\pi}{M}\right)  \ \left(  \cos s+s\sin s\right)
=0
\end{equation}
which has the same form as the circular case with the replacements
$M\rightarrow\frac{M}{2}$ and $n\rightarrow\frac{n+1}{2}$. $D_{1}$ and $D_{2}$
may be written out in full as
\begin{equation}
D_{1}=\frac{M}{n\,\pi}\left(
\begin{array}
[c]{c}%
\int_{0}^{\frac{2\pi}{M}-s}d\theta\left(  1+\left\|  \mathbf{n}\left(
\theta\right)  -r\,\mathbf{n}\left(  0\right)  \right\|  ^{2}\right) \\
+\int_{\frac{2\pi}{M}-s}^{\frac{2\pi}{M}}d\theta\,f\left(  \theta\right)
\left(  1+\left\|  \mathbf{n}\left(  \theta\right)  -r\,\mathbf{n}\left(
0\right)  \right\|  ^{2}\right) \\
+\int_{\frac{2\pi}{M}-s}^{\frac{2\pi}{M}}d\theta\,f\left(  \frac{4\pi}%
{M}-\theta\right)  \left(  1+\left\|  \mathbf{n}\left(  \theta\right)
-r\,\mathbf{n}\left(  \frac{4\pi}{M}\right)  \right\|  ^{2}\right)
\end{array}
\right)
\end{equation}%
\begin{equation}
D_{2}=\frac{2\left(  n-1\right)  M}{n\,\pi}\left(
\begin{array}
[c]{c}%
\int_{0}^{\frac{2\pi}{M}-s}d\theta\left\|  \mathbf{n}\left(  \theta\right)
-\frac{1}{2}r\,\mathbf{n}\left(  0\right)  \right\|  ^{2}\\
+\int_{\frac{2\pi}{M}-s}^{\frac{2\pi}{M}}d\theta\left\|
\begin{array}
[c]{c}%
\mathbf{n}\left(  \theta\right)  -\frac{1}{2}r\,f\left(  \theta\right)
\,\mathbf{n}\left(  0\right) \\
-\frac{1}{2}r\,f\left(  \frac{4\pi}{M}-\theta\right)  \mathbf{n}\left(
\frac{4\pi}{M}\right)
\end{array}
\right\|  ^{2}%
\end{array}
\right)
\end{equation}
The optimum $f\left(  \theta\right)  $ and $r$ may be substituted into
$D_{1}+D_{2}$, the integrations evaluated, and then the condition that the
optimum $s$ must satisfy may be used to simplify the result, to yield the
minimum $D_{1}+D_{2}$ as
\begin{equation}
D_{1}+D_{2}=4-\frac{n}{n-1}\frac{M}{2\pi}\left(  2s+\sin\left(  2s\right)
\right)
\end{equation}
which has the same form as the circular case plus an extra contribution of 2.

\subsection{Toroidal Manifold: 3 Overlapping Posterior Probabilities}

\label{Appendix:ToroidalManifold3}For $\frac{\pi}{M}\leq s\leq\frac{2\pi}{M}$
the functional form of $p\left(  \theta\right)  $ may be obtained directly
from the circular case with the replacement $M\rightarrow\frac{M}{2}$, so
that
\begin{equation}
p\left(  \theta\right)  =\left\{
\begin{array}
[c]{ll}%
f_{1}\left(  \theta\right)  & 0\leq\left|  \theta\right|  \leq-\frac{2\pi}%
{M}+s\\
f_{2}\left(  \theta\right)  & -\frac{2\pi}{M}+s\leq\left|  \theta\right|
\leq\frac{6\pi}{M}-s\\
f_{3}\left(  \theta\right)  & \frac{6\pi}{M}-s\leq\left|  \theta\right|
\leq\frac{2\pi}{M}+s\\
0 & \left|  \theta\right|  \geq\frac{2\pi}{M}+s
\end{array}
\right.
\end{equation}%
\begin{align}
f_{1}\left(  \theta\right)    =&\frac{1}{2}\cos\left(  \theta\right)
\sec\left(  \frac{2\pi}{M}-s\right) \nonumber\\
& +a_{1}\left(  1-\cos\left(  \theta\right)  \ \sec\left(  \frac{2\pi}%
{M}-s\right)  \right) \nonumber\\
& +b_{2}\cos\left(  \theta\right)  \csc\left(  \frac{2\pi}{M}\right)
\sin\left(  \frac{4\pi}{M}-s\right)  \sec\left(  \frac{2\pi}{M}-s\right)
\nonumber\\
f_{2}\left(  \theta\right)    =&\frac{1}{2}+b_{2}\ \left(  \cos\left(
\theta\right)  -\cot\left(  \frac{2\pi}{M}\right)  \ \sin\left(
\theta\right)  \right) \nonumber\\
f_{3}\left(  \theta\right)    =&\frac{1}{2}\left(  1-\csc\left(  \frac{4\pi
}{M}-2\ s\right)  \sin\left(  \frac{6\pi}{M}-s-\theta\right)  \right)
\nonumber\\
& +\frac{1}{2}a_{1}\left(  \cos\left(  \frac{4\pi}{M}-\theta\right)
\sec\left(  \frac{2\pi}{M}-s\right)  -1\right) \nonumber\\
& +b_{2}\csc\left(  \frac{2\pi}{M}\right)  \csc\left(  \frac{4\pi}%
{M}-2\ s\right)  \sin\left(  \frac{4\pi}{M}-s\right)  \sin\left(  \frac{2\pi
}{M}+s-\theta\right)
\end{align}
$D_{1}+D_{2}$ must be stationary w.r.t. variation of $p\left(  \theta\right)
$ in each of the 3 intervals $0\leq\theta\leq-\frac{2\pi}{M}+s$ (interval 1),
$-\frac{2\pi}{M}+s\leq\theta\leq\frac{6\pi}{M}-s$ (interval 2), and
$\frac{6\pi}{M}-s\leq\theta\leq\frac{2\pi}{M}+s$ (interval 3). The coefficient
of the $\cos2\theta$ term in interval 2 yields
\begin{equation}
\ b_{2}\,r\,\left(  n+b_{2}\,r-b_{2}r\,n\right)  \cos\left(  \frac{4\pi}%
{M}\right)  =0
\end{equation}
which has the same form as the circular case with the replacements
$M\rightarrow\frac{M}{2}$ and $n\rightarrow\frac{2n}{n+1}$, which has the
solution
\begin{equation}
b_{2}=\frac{n}{\left(  n-1\right)  \,r}%
\end{equation}
which may be substituted back into the coefficient of the $\cos\theta$ term in
interval 1 to yield
\begin{align}
0   =&r\sec\left(  \frac{2\pi}{M}-s\right)  \sin\left(  \frac{2\pi}{M}\right)
\nonumber\\
& \times\left(
\begin{array}
[c]{c}%
\left(  n-1\right)  \,\left(  -6\,a_{1}^{2}+7\,a_{1}-2\right)  \,r\sin\left(
\frac{2\pi}{M}\right) \\
+\left(  n-1\right)  \,\left(  2\,a_{1}^{2}-3\,a_{1}+1\right)  \,r\sin\left(
\frac{6\pi}{M}\right) \\
+2n\,\left(  a_{1}\sin\left(  \frac{4\pi}{M}-s\right)  +\left(  1-a_{1}%
\right)  \sin\left(  \frac{8\pi}{M}-s\right)  \right)
\end{array}
\right)
\end{align}
and also substituted back into the coefficient of the $\sin\theta$ term in
interval 3 to yield
\begin{align}
0   =&r\,\cos\left(  \frac{2\pi}{M}\right)  \,\csc\left(  \frac{2\pi}%
{M}-s\right)  \,\sec\left(  \frac{2\pi}{M}-s\right)  \,\sin^{2}\left(
\frac{2\pi}{M}\right)  \nonumber\\
& \times\left(
\begin{array}
[c]{c}%
-\left(  n-1\right)  \,r\,\left(
\begin{array}
[c]{c}%
-2\,a_{1}\left(  3\,a_{1}-2\right)  \cos\left(  \frac{4\pi}{M}-s\right) \\
-2\,\left(  a_{1}-1\right)  \,\,a_{1}\cos\left(  \frac{4\,\pi}{M}+s\right) \\
+\left(  1-2a_{1}+2\,a_{1}^{2}\right)  \,\cos\left(  \frac{8\,\pi}{M}-s\right)
\\
+\left(  1-4\,a_{1}+6\,a_{1}^{2}\right)  \,\cos\left(  s\right)
\end{array}
\right) \\
+2n\,\left(
\begin{array}
[c]{c}%
\left(  a_{1}+1\right)  \cos\left(  \frac{2\pi}{M}\right)  -\left(
a_{1}-1\right)  \cos\left(  \frac{6\,\pi}{M}\right) \\
-2\,a_{1}\sin\left(  \frac{2\pi}{M}\right)  \sin\left(  \frac{8\pi}%
{M}-2\,s\right)
\end{array}
\right)
\end{array}
\right)
\end{align}
both of which have the same form as the circular case with the replacements
$M\rightarrow\frac{M}{2}$ and $n\rightarrow\frac{2n}{n+1}$. These two
conditions may be solved for $a_{1}$ and $r$ to yield
\begin{equation}
a_{1}=\frac{\cos\left(  \frac{4\pi}{M}\right)  }{\cos\left(  \frac{4\pi}%
{M}\right)  -1}%
\end{equation}
and
\begin{equation}
r=\frac{2n}{n-1}\cos\left(  \frac{4\pi}{M}-s\right)  \,\sec\left(  \frac{2\pi
}{M}\right)
\end{equation}
The solutions for $a_{1}$ and $b_{2}$ may be substituted back into the
expressions for the $f_{i}\left(  \theta\right)  $ to reduce them to the form
\begin{align}
f_{1}\left(  \theta\right)   & =-\frac{1}{4}\left(  \cos\left(  \frac{8\pi}%
{M}-s\right)  +\cos s-2\cos\left(  \frac{2\pi}{M}\right)  \cos\theta\right)
\csc^{2}\left(  \frac{2\pi}{M}\right)  \sec\left(  \frac{4\pi}{M}-s\right)
\nonumber\\
f_{2}\left(  \theta\right)   & =\frac{1}{2}\left(  \cot\left(  \frac{2\pi}%
{M}\right)  \sec\left(  \frac{4\pi}{M}-s\right)  \sin\left(  \frac{2\pi}%
{M}-\theta\right)  +1\right) \nonumber\\
f_{3}\left(  \theta\right)   & =-\frac{1}{4}\csc^{2}\left(  \frac{2\pi}%
{M}\right)  \left(  \cos\left(  \frac{6\pi}{M}-\theta\right)  \sec\left(
\frac{6\pi}{M}-s\right)  -1\right)
\end{align}
which have the same form as the circular case with the replacement
$M\rightarrow\frac{M}{2}$. $D_{1}+D_{2}$ must be stationary w.r.t. variation
of $r$. This yields a transcendental equation that must be satisfied by the
optimum solution for $s$ as
\begin{equation}
\frac{1}{n}\frac{\cos\left(  \frac{4\pi}{M}-s\right)  }{\cos\left(  \frac
{2\pi}{M}\right)  }-\frac{n-1}{2n}\frac{M}{2\pi}\cos\left(  \frac{2\pi}%
{M}\right)  \left(  \sin\left(  \frac{4\pi}{M}-s\right)  -\left(  \frac{4\pi
}{M}-s\right)  \ \cos\left(  \frac{4\pi}{M}-s\right)  \right)  =0
\end{equation}
which has the same form as the circular case with the replacements
$M\rightarrow\frac{M}{2}$ and $n\rightarrow\frac{n+1}{2}$. $D_{1}$ and $D_{2}$
may be written out in full as
\begin{equation}
D_{1}=\frac{M}{2\,n\,\pi}\left(
\begin{array}
[c]{c}%
\int_{0}^{-\frac{2\pi}{M}+s}d\theta\,\left(
\begin{array}
[c]{c}%
f_{1}\left(  \theta\right)  \left(  1+\left\|  \mathbf{n}\left(
\theta\right)  -r\,\mathbf{n}\left(  0\right)  \right\|  ^{2}\right) \\
+f_{3}\left(  \frac{4\pi}{M}-\theta\right)  \left(  1+\left\|  \mathbf{n}%
\left(  \theta\right)  -r\,\mathbf{n}\left(  \frac{4\pi}{M}\right)  \right\|
^{2}\right) \\
+f_{3}\left(  \frac{4\pi}{M}+\theta\right)  \left(  1+\left\|  \mathbf{n}%
\left(  \theta\right)  -r\,\mathbf{n}\left(  -\frac{4\pi}{M}\right)  \right\|
^{2}\right)
\end{array}
\right) \\
+\int_{-\frac{2\pi}{M}+s}^{\frac{6\pi}{M}-s}d\theta\,\left(
\begin{array}
[c]{c}%
f_{2}\left(  \theta\right)  \left(  1+\left\|  \mathbf{n}\left(
\theta\right)  -r\,\mathbf{n}\left(  0\right)  \right\|  ^{2}\right) \\
+f_{2}\left(  \frac{4\pi}{M}-\theta\right)  \left(  1+\left\|  \mathbf{n}%
\left(  \theta\right)  -r\,\mathbf{n}\left(  \frac{4\pi}{M}\right)  \right\|
^{2}\right)
\end{array}
\right) \\
+\int_{\frac{6\pi}{M}-s}^{\frac{4\pi}{M}}d\theta\,\left(
\begin{array}
[c]{c}%
f_{3}\left(  \theta\right)  \left(  1+\left\|  \mathbf{n}\left(
\theta\right)  -r\,\mathbf{n}\left(  0\right)  \right\|  ^{2}\right) \\
+f_{1}\left(  \frac{4\pi}{M}-\theta\right)  \left(  1+\left\|  \mathbf{n}%
\left(  \theta\right)  -r\,\mathbf{n}\left(  \frac{4\pi}{M}\right)  \right\|
^{2}\right) \\
+f_{3}\left(  \frac{8\pi}{M}-\theta\right)  \left(  1+\left\|  \mathbf{n}%
\left(  \theta\right)  -r\,\mathbf{n}\left(  \frac{8\pi}{M}\right)  \right\|
^{2}\right)
\end{array}
\right)
\end{array}
\right)
\end{equation}%
\begin{equation}
D_{2}=\frac{\left(  n-1\right)  M}{n\,\pi}\left(
\begin{array}
[c]{c}%
\int_{0}^{-\frac{2\pi}{M}+s}d\theta\left\|
\begin{array}
[c]{c}%
\mathbf{n}\left(  \theta\right)  -\frac{1}{2}f_{1}\left(  \theta\right)
\,r\,\mathbf{n}\left(  0\right) \\
-\frac{1}{2}f_{3}\left(  \frac{4\pi}{M}-\theta\right)  \,r\,\mathbf{n}\left(
\frac{4\pi}{M}\right) \\
-\frac{1}{2}f_{3}\left(  \frac{4\pi}{M}+\theta\right)  \,r\,\mathbf{n}\left(
-\frac{4\pi}{M}\right)
\end{array}
\right\|  ^{2}\\
+\int_{-\frac{2\pi}{M}+s}^{\frac{6\pi}{M}-s}d\theta\,\left\|
\begin{array}
[c]{c}%
\mathbf{n}\left(  \theta\right)  -\frac{1}{2}f_{2}\left(  \theta\right)
\,r\,\mathbf{n}\left(  0\right) \\
-\frac{1}{2}f_{2}\left(  \frac{4\pi}{M}-\theta\right)  \,r\,\mathbf{n}\left(
\frac{4\pi}{M}\right)
\end{array}
\right\|  ^{2}\\
+\int_{\frac{6\pi}{M}-s}^{\frac{4\pi}{M}}d\theta\,\left\|
\begin{array}
[c]{c}%
\mathbf{n}\left(  \theta\right)  -\frac{1}{2}f_{3}\left(  \theta\right)
\,r\,\mathbf{n}\left(  0\right) \\
-\frac{1}{2}f_{1}\left(  \frac{4\pi}{M}-\theta\right)  \,r\,\mathbf{n}\left(
\frac{4\pi}{M}\right) \\
-\frac{1}{2}f_{3}\left(  \frac{8\pi}{M}-\theta\right)  \,r\,\mathbf{n}\left(
\frac{8\pi}{M}\right)
\end{array}
\right\|  ^{2}%
\end{array}
\right)
\end{equation}
The optimum $f_{i}\left(  \theta\right)  $ and $r$ may be substituted into
$D_{1}+D_{2}$, the integrations evaluated, and then the condition that the
optimum $s$ must satisfy may be used to simplify the result, to yield the
minimum $D_{1}+D_{2}$ as
\begin{align}
D_{1}+D_{2}   =&\frac{n\left(  \left(  n-1\right)  \left(  2\frac{n-2}%
{n}\ -\frac{M}{2\pi}\,s\right)  -2\sec^{2}\left(  \frac{2\pi}{M}\right)
\right)  }{\left(  n-1\right)  ^{2}}\nonumber\\
& -\frac{n\left(  \left(  n-1\right)  \ \left(  2-\frac{M}{2\pi}\,s\right)
+2\sec^{2}\left(  \frac{2\pi}{M}\right)  \right)  }{\left(  n-1\right)  ^{2}%
}\cos\left(  \frac{8\pi}{M}-2s\right)
\end{align}

\end{document}